\documentclass[journal]{IEEEtran}
\usepackage{float}
\usepackage{booktabs}
\usepackage{graphicx}
\usepackage{subfig}
\usepackage{amssymb}
\usepackage{enumitem,amssymb}
\usepackage{bbding} 
\usepackage{cite}
\usepackage{indentfirst}
\usepackage{enumerate}
\usepackage{hyperref}
\usepackage{multirow}
\usepackage{makecell}  
\usepackage{nicematrix}
\usepackage[ruled]{algorithm2e}

\makeatletter
\newcommand{\removelatexerror}{\let\@latex@error\@gobble}
\makeatother
\makeatletter
\renewcommand{\maketag@@@}[1]{\hbox{\m@th\normalsize\normalfont#1}}%
\makeatother

\hypersetup{
	colorlinks=true,
	linkcolor=red,
	urlcolor=blue
}

\usepackage{tikz,xcolor}
\definecolor{lime}{HTML}{A6CE39}
\DeclareRobustCommand{\orcidicon}{%
	\begin{tikzpicture}
	\draw[lime, fill=lime] (0,0) 
	circle [radius=0.134] 
	node[white] {{\fontfamily{qag}\selectfont \tiny ID}};    \draw[white, fill=white] (-0.0625,0.095) 
	circle [radius=0.007];    \end{tikzpicture}
	\hspace{-2mm}}
\foreach \x in {A, ..., Z}{%
	\expandafter\xdef\csname orcid\x\endcsname{\noexpand\href{https://orcid.org/\csname orcidauthor\x\endcsname}{\noexpand\orcidicon}}
}

\begin{document}
	
\captionsetup[figure]{name={Fig.},labelsep=period}
\captionsetup[table]{name={TABLE},labelsep=period}

\title{Local-Global Temporal Difference Learning for Satellite Video Super-Resolution}

\author{Yi~Xiao,
       Qiangqiang~Yuan\orcidA{}, ~\IEEEmembership{Member,~IEEE,}
       Kui~Jiang, ~\IEEEmembership{Member,~IEEE,}
       Xianyu~Jin, 
       Jiang~He,
       Liangpei~Zhang\orcidC{}, ~\IEEEmembership{Fellow,~IEEE}, 
       and~Chia-wen~Lin\orcidB{}, ~\IEEEmembership{Fellow,~IEEE.}
        
\thanks{This work was supported  in part by the National Natural
	Science Foundation of China under Grant 42230108 and 61971319. (\emph{Corresponding author: Qiangqiang Yuan and Kui Jiang}.)}
\thanks{Yi Xiao is with the School of Geodesy and Geomatics, Wuhan University, Wuhan 430079, China (e-mail: xiao\_yi@whu.edu.cn).}
\thanks{Qiangqiang Yuan is with the School of Geodesy and Geomatics, Wuhan University (e-mail: yqiang86@gmail.com).}
\thanks{Kui Jiang is with the Faculty of Computing, Harbin Institute of Technology (e-mail: kuijiang\_1994@163.com).}
\thanks{Xianyu Jin is with the School of Geodesy and Geomatics, Wuhan University (e-mail: jin\_xy@whu.edu.cn).}
\thanks{Jiang He is with the School of Geodesy and Geomatics, Wuhan University (e-mail: jiang\_he@whu.edu.cn).}
\thanks{Liangpei Zhang is with the State Key Laboratory of Information Engineering in Surveying, Mapping, and Remote Sensing, Wuhan University (e-mail: zlp62@whu.edu.cn).}
\thanks{Chia-Wen Lin is with the Department of Electrical Engineering and the Institute of Communications Engineering, National Tsing Hua University (e-mail: cwlin@ee.nthu.edu.tw).}}

%
%

\markboth{Submitted to IEEE Transactions on Circuits and Systems for Video Technology}%
{Shell \MakeLowercase{\textit{et al.}}: Bare Demo of IEEEtran.cls for IEEE Journals}
%



\maketitle

\begin{abstract}
Optical-flow-based and kernel-based approaches have been extensively explored for temporal compensation in satellite Video Super-Resolution (VSR). However, these techniques are less generalized in large-scale or complex scenarios, especially in satellite videos. In this paper, we propose to exploit the well-defined temporal difference for efficient and effective temporal compensation. To fully utilize the local and global temporal information within frames, we systematically modeled the short-term and long-term temporal discrepancies since we observe that these discrepancies offer distinct and mutually complementary properties. Specifically, we devise a Short-term Temporal Difference Module (S-TDM) to extract local motion representations from RGB difference maps between adjacent frames, which yields more clues for accurate texture representation. To explore the global dependency in the entire frame sequence, a Long-term Temporal Difference Module (L-TDM) is proposed, where the differences between forward and backward segments are incorporated and activated to guide the modulation of the temporal feature, leading to a holistic global compensation. Moreover, we further propose a Difference Compensation Unit (DCU) to enrich the interaction between the spatial distribution of the target frame and temporal compensated results, which helps maintain spatial consistency while refining the features to avoid misalignment. Rigorous objective and subjective evaluations conducted across five mainstream video satellites demonstrate that our method performs favorably against state-of-the-art approaches. Code will be available at \url{https://github.com/XY-boy/LGTD}
\end{abstract}

\begin{IEEEkeywords}
Satellite video, super-resolution, temporal difference, local-global compensation, remote sensing.
\end{IEEEkeywords}

%
\IEEEpeerreviewmaketitle

\section{Introduction}

\IEEEPARstart{C}{ompared} to traditional static remote sensing images, video satellite provides continuous information to a specific area, which is crucial for dynamic earth observation. Therefore, it has been widely applied to dynamic scene applications such as change detection \cite{zhou2022spatial}, object tracking \cite{zhao2022satsot}, and traffic monitoring \cite{he2021inferring}. However, the spatial resolution of satellite video is usually contaminated by the complex aerial environment, limited by the intrinsic resolution of satellite video sensors ($\sim$1m), and degraded by data compression. Consequently, high-frequency information in the satellite video may be lost, which dramatically reduces the visual quality and degrades performance in subsequent applications. To this end, it is essential to improve the spatial resolution of satellite videos.

Compared to upgrading the hardware to promote the resolution, software-based super-resolution (SR) technologies greatly save maintenance, transmission, and storage costs \cite{hj1, he2023spectral, jk1, 10144690, zq1}. Early SR approaches usually use hand-craft priors to make this problem well-posed. However, their performance is limited by these laborious priors, and they often suffer from complex optimization problems. By contrast, convolution neural networks (CNNs) \cite{niu2022ms2net, 9792407,deepcams, wy1, wy2, liu2023efficient} have emerged as a preferable choice due to their powerful non-linear representation ability. In spite of achieving a decent result in single image SR (SISR), these algorithms reach a bottleneck as the temporal information is not considered \cite{liu2022temporal, sun2021fpga,jxy,zhang2020multi, zhang2022optical, yi2019multi}. Furthermore, most video SR (VSR) methods are specialized for natural videos and are not easily applied to satellite videos. Hence, more efforts should be dedicated to developing an applicable method for satellite VSR.


The key to the success of VSR is compensating the misaligned pixels using temporal redundancy from neighboring frames. Most previous works are engaged in optical flow \cite{haris2019recurrent, wang2020deep, xiao2022space} and kernel estimation \cite{yu2022memory, wang2019edvr} for temporal compensation. The optical-flow-based methods often employ an extra component to get the flow maps between frames or jointly optimize the flow estimation sub-network with the whole network. However, flow estimation is a laborious task and would introduce high complexity. The kernel-based methods, \emph{e.g.}, 3D convolution \cite{isobe2020video, jo2018deep} and deformable convolution (DConv) \cite{9530280}, either ignore the valuable temporal priors or have a limited spatial-temporal receptive field. Moreover, these approaches are easily collapsed in complex motion. Recently, several works \cite{chan2021basicvsr, zhu2022fffn} proposed to propagate rich historical information recurrently. Nevertheless, they acquire a considerable memory cost to cache future and past frames and neglect the potential of local compensation because the recurrent structure naturally focuses on long-range dependencies.

\begin{figure}[!t]
\centering
\includegraphics[width=3.5in]{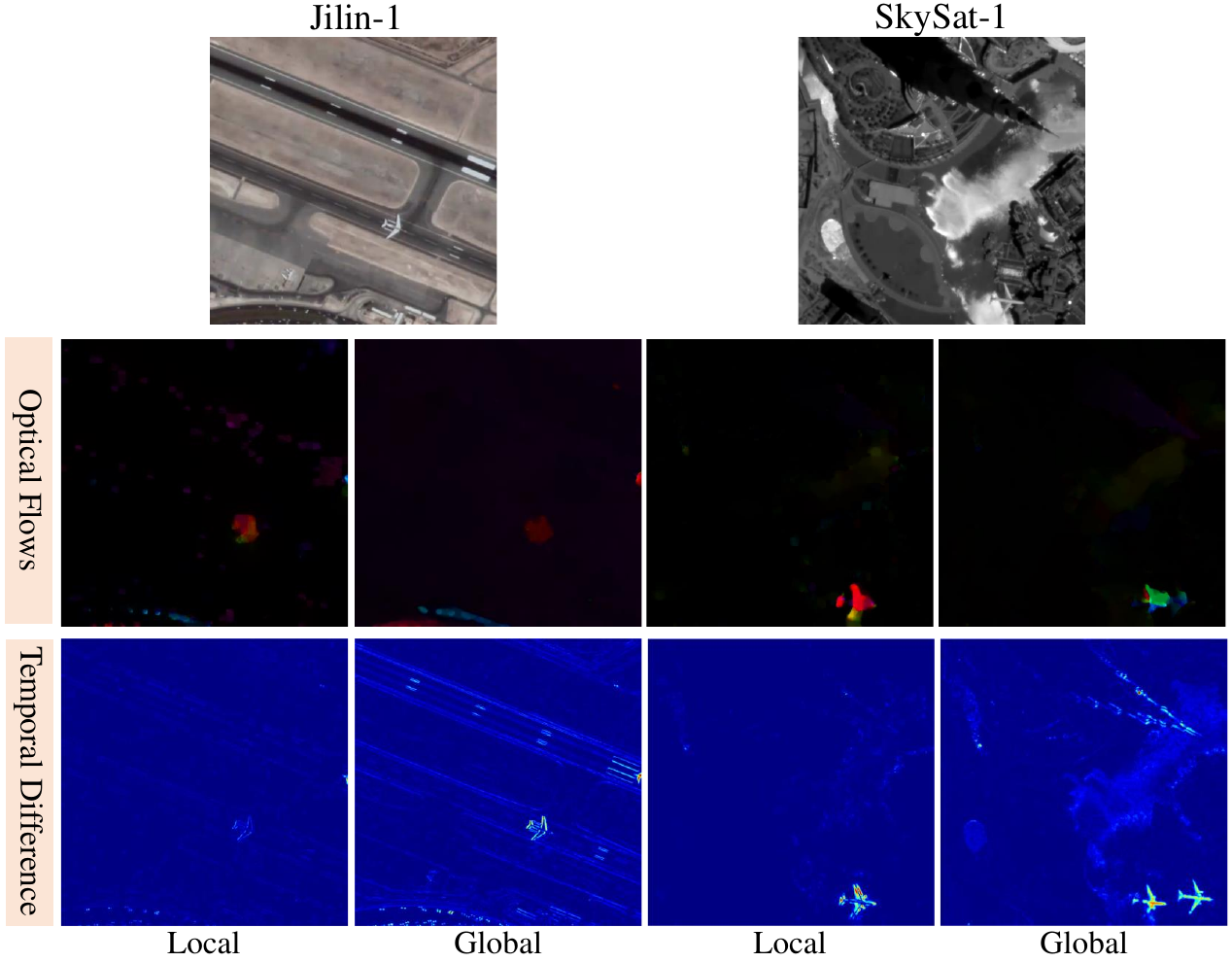}%
\captionsetup{font={scriptsize}}   
\caption{Comparison between RGB temporal differences and optical flow maps. The optical flow maps were generated by PyFlow \cite{liu2009beyond}. Temporal differences activate more accurate and sharp cues than optical flows. Besides, the local temporal difference and global temporal difference are not equally informative as they reflect a different level of difference.}
\label{diff}
\end{figure}

In contrast to natural video, satellite video exhibits unique characteristics that make temporal compensation more complicated. Since the static background is the main present content of remote-sensing scenes, both optical flow estimation and deformable convolution inevitably involve redundant calculations in the background pixels. In particular, optical flow estimation is fragile to the multi-scale small objects and complex boundary tremors that commonly appear in satellite videos. These inherent challenges frequently lead to inaccuracies in the temporal compensation process. Recently, the temporal difference is demonstrated informative in modeling motion information and can be an alternative to optical flow \cite{zhao2018recognize, isobe2022look, wang2021tdn}. Inspired by these successes, we make an attempt to apply the explicit temporal difference to satellite VSR tasks. As shown in Fig. \ref{diff}, the motion information could be well-activated in temporal difference maps, which implied more accurate and sharp texture information than optical flows. Additionally, temporal difference information of remote-sensing images is sparse because the massive redundancy (highly similar background across frames) is reduced to low values after subtraction. As a result, these temporal differences can serve as sparse signals, facilitating efficient inferences in large-scale remote-sensing imagery. Furthermore, both the local and global contents are crucial for compensation in the sense that they are able to capture the distinctive and complementary properties of motion information. As illustrated in Fig. \ref{diff}, the local temporal difference is prominent on the boundary of moving objects, while the global temporal difference contains more edge and shape details. This observation drives us to introduce a distinctive two-level temporal modeling strategy, systematically addressing both local and global temporal compensation.

Specifically, this paper proposed a Local-Global Temporal Difference learning network (LGTD) for satellite VSR. 
The proposed Short-term Temporal Difference Module (S-TDM) derives the local information from adjacent frames with low variance, while the global information with large variance is inferred by a Long-term Temporal Difference Module (L-TDM). Instead of decomposing neighboring frames into high-variance and low-variance components using laborious pre-processing \cite{isobe2022look}, our LGTD directly learns complementary information from RGB differences. In S-TDM, the RGB difference maps are transformed into local motion representations and supplied to the target frame with lateral connections. While some previous efforts \cite{isobe2022look} have utilized recurrent structures to investigate forward and backward temporal differences, our L-TDM emphasizes activating a holistic motion representation from entire frame sequences, resulting in a robust global compensation. To alleviate the potential misalignment caused by temporal difference learning, we enhance the interaction between the target frame and temporal information yielded from TDMs through a Difference Compensation Unit (DCU). In this manner, our S-TDM and L-TDM could focus on valuable information to reconstruct the target frame and maintain spatial consistency. In short, this paper contributes as follows:
\begin{description}
\item 1) Different from previous temporal compensation approaches, we generalize the idea of temporal difference to achieve temporal compensation for satellite VSR. The proposed local-global temporal difference learning is computation-friendly and could provide an alternative to optical flows. 
\item 2) Our S-TDM and L-TDM could systematically utilize both short-term and long-term temporal complementary information from local and global motion patterns.
\item 3) We devise a Difference Compensation Unit (DCU) to alleviate misalignment in temporal difference learning and to help TMDs focus on capturing temporal information that is beneficial to satellite VSR.
\item 4) Compared with optical-flow-based and kernel-based approaches, our method achieves favorable quantitative and qualitative results on five mainstream video satellites.
\end{description}

The remainder of this paper is organized as follows: Section \ref{related} reviews video super-resolution, Section \ref{meth} involves details of the proposed method, Section \ref{exp} contains experiments and analysis, and Section \ref{conclu} is the conclusion.

\section{Related Work}\label{related}
\subsection{Video Super-Resolution}
Here, we first review optical-flow-based VSR methods that employ optical flow to describe motion information, followed by explicit motion compensation. Then, we present kernel-based implicit alignment methods. Finally, we introduce the recurrent propagation VSR framework.

\subsubsection{Optical-flow-based VSR methods}Generally, the flow estimation algorithms can be divided into traditional and deep-learning-based approaches. In the traditional approach, Deep-DE \cite{liao2015video} used ${\ell _1}$ flow \cite{brox2004high} to generate a series of SR drafts, then the bilinear upsampled LR target frame is concatenated with these drafts and sent to CNN and deconvolution layer for reconstruction. VSRNet \cite{kappeler2016video} adopted the Druleas algorithm \cite{drulea2011total} to compute the flow maps and proposed a filter symmetry mechanism to compensate neighboring frames. Recently,  Haris \emph{et al.} \cite{haris2019recurrent} introduced a recurrent back-projection network (RBPN), employing an external package named PyFlow \cite{liu2009beyond} for flow estimation. On the other hand, deep learning-based flow estimation often relies on CNNs to predict flows. For instance, Caballero \emph{et al.} \cite{caballero2017real} introduced a motion compensation transformer (MCT) that simultaneously learns motion information and performs motion compensation. Wang \emph{et al.} \cite{wang2020deep} designed a unified framework to jointly super-resolve optical flow maps and target frames, addressing the resolution gap between low-resolution flow maps and latent high-resolution features. Additionally, more elaborate networks like FlowNet \cite{ilg2017flownet} and SpyNet \cite{ranjan2017optical} are commonly employed CNN architectures for optical flow estimation.

\par However, both traditional and deep learning-based methods may not guarantee the accuracy of flow estimation, particularly in large-scale remote sensing scenarios, leading to several performance drops. Besides, incorporating external algorithms and specific sub-networks for flow estimation will inevitably increase computation consumption. 

\subsubsection{Kernel-based VSR methods}This kind of approach usually implicitly parameterizes the temporal compensation into convolution kernels, with notable efforts in 3D convolution \cite{kim2019video}, non-local module \cite{yi2020progressive}, deformable convolution (DConv) \cite{wang2019edvr}, and Transformer \cite{liu2022learning, shi2022rethinking}.
Jo \emph{et al.} \cite{jo2018deep} learned a 3D Dynamic Upsampling Filter (DUF) for each pixel to avoid explicit motion estimation and compensation. Similarly, Wen \emph{et al.} \cite{9700735} employed spatio-temporal adaptive filters to achieve implicit alignment. However, 3D convolution is not sophisticated in modeling the temporal priors and would increase the complexity. Non-local-based approaches hire the non-local attention mechanism to capture long-range dependencies, which increase the receptive field of temporal information. Yi \emph{et al.} \cite{yi2020progressive} proposed a progressive fusion strategy to aggregate the non-local spatial-temporal information. Yu \emph{et al.} \cite{yu2022memory} established a novel Memory-Augment Non-local Attention (MANA) to memory details cross-frames. Nevertheless, non-local learning introduces significant inference complexity and yields limited improvement. The DConv is first proposed in \cite{dai2017deformable} and carried forward by many works like TDAN \cite{tian2020tdan}, EDVR \cite{wang2019edvr}, and D3DNet \cite{ying2020deformable}. The key to DConv is attaining and applying the offset parameters on the convolution grid to yield a deformable sampling position. Hence, the convolution can involve motion information inside the receptive field. TDAN \cite{tian2020tdan} used shallow convolution layers to predict the offsets. EDVR \cite{wang2019edvr} enlarged the receptive field with the help of a pyramid structure and coarse-to-fine alignment. D3DNet introduced DConv to 3D dimension for a larger temporal receptive field. More recently, Isobe \emph{et al.} \cite{isobe2022look} developed a temporal difference modeling network (ETDM) to achieve alignment by computing the temporal difference in two subsets.
\par Although kernel-based methods bring promising improvements, they still face harsh convergence conditions and inefficient computation.

\subsubsection{Recurrent propagation VSR methods}Such methods aim to fully utilize the long-range dependencies by propagating rich future or past information in a recurrent manner \cite{Jeelani_2023_CVPR}. For example, Huang \emph{et al.} \cite{huang2017video} proposed a recurrent bidirectional network with forward and backward sub-networks. The two similar sub-networks are responsible for learning the previous and past frames. In \cite{tao2017detail, guo2017building, zhu2019residual}, the authors employ LSTM to leverage the long-distance temporal properties. While the RNN-based network is suitable for spatial-temporal modeling, the errors caused by misalignment may accumulate with the input length increase. Toward this end, Chan \emph{et al.} \cite{chan2021basicvsr} come up with an information-refill and propagation strategy to ease the misalignment accumulation. In addition, they enhance the interaction of bi-directional temporal information by forward and backward propagation. Recently, BasicVSR++ \cite{chan2022basicvsr++} is proposed to strengthen the alignment and propagation process. More recently, Chiche \emph{et al.} \cite{chiche2022stable} proposed a stable recurrent network for long-term video super-resolution. Xie \emph{et al.} \cite{xie2023mitigating} proposed mitigating the hidden state artifacts to generate sharp details for recurrent propagation.

\par To sum up, recurrent propagation VSR methods can benefit from RNNs or LSTMs to capture the long-range dependency with a lightweight design. However, it is hard to train an RNN due to the gradient vanishing. Furthermore, they still lack using short-term information between adjacent frames since recurrent structures naturally focus on long-term information.

\subsection{Satellite Video Super-resolution}
In satellite VSR, early efforts \cite{luo2017video} often transfer the single-image super-resolution (SISR) frameworks on satellite video without specific consideration of the characteristics of remote sensing imagery. Later, more methods applicable to remote sensing images were explored \cite{xy2}. Jiang \emph{et al.} \cite{8677274} proposed a GAN-based framework to enhance the high-frequency edge information. Zhang \emph{et al.} \cite{zhang2020scene} take account for scene variation in remote sensing images and put forward a scene-adaptive strategy to alleviate the performance drop across scenes. He \emph{et al.} \cite{he2020unified} proposed a hybrid-scale network to fully extracted the multi-scale information. However, these methods are SISR, thus failing to model the temporal information and having a bottleneck. Recently, Liu \emph{et al.} \cite{liu2020satellite} designed a traditional method, which adaptively examined the non-local similarity in wide-range remote sensing imagery. He \emph{et al.} \cite{he2021multiframe} proposed to employ 3D convolution to implicitly achieve motion compensation. But it increases computational consumption and has gained limited performance without sophisticated design. More recently, Xiao \emph{et al.} \cite{9530280} proposed a multi-scale DConv model for precise alignment and designed a temporal grouping projection to realize effective spatial-temporal fusion. They further put forward to employ deformable attention for temporal alignment \cite{10172076}.
\par Years of effort have witnessed remarkable progress in satellite VSR, but we still need an efficient and straightforward solution to excavate the high-redundant temporal information in satellite video. Unlike the TDN \cite{wang2021tdn} that primarily focuses on modeling temporal differences for efficient action recognition, our method addresses the challenges specific to satellite video super-resolution, where motion is often scarce. Inspired by TDN, we exploit temporal difference learning to provide efficient yet effective motion cues. In particular, while TDN directly captures motion dynamics by explicitly modeling RGB-wise differences across video sequences, our approach first performs feature-level alignment using deformable convolutions. This helps reduce misalignment caused by large motions and mitigate the difficulties encountered in subsequent temporal difference learning, better handling the sparse temporal variations characteristic of satellite videos. These modifications enable our method to achieve more robust alignment performance, which in turn facilitates superior reconstruction quality in satellite VSR tasks.

\begin{figure*}[!t]
\centering
\includegraphics[width=7in]{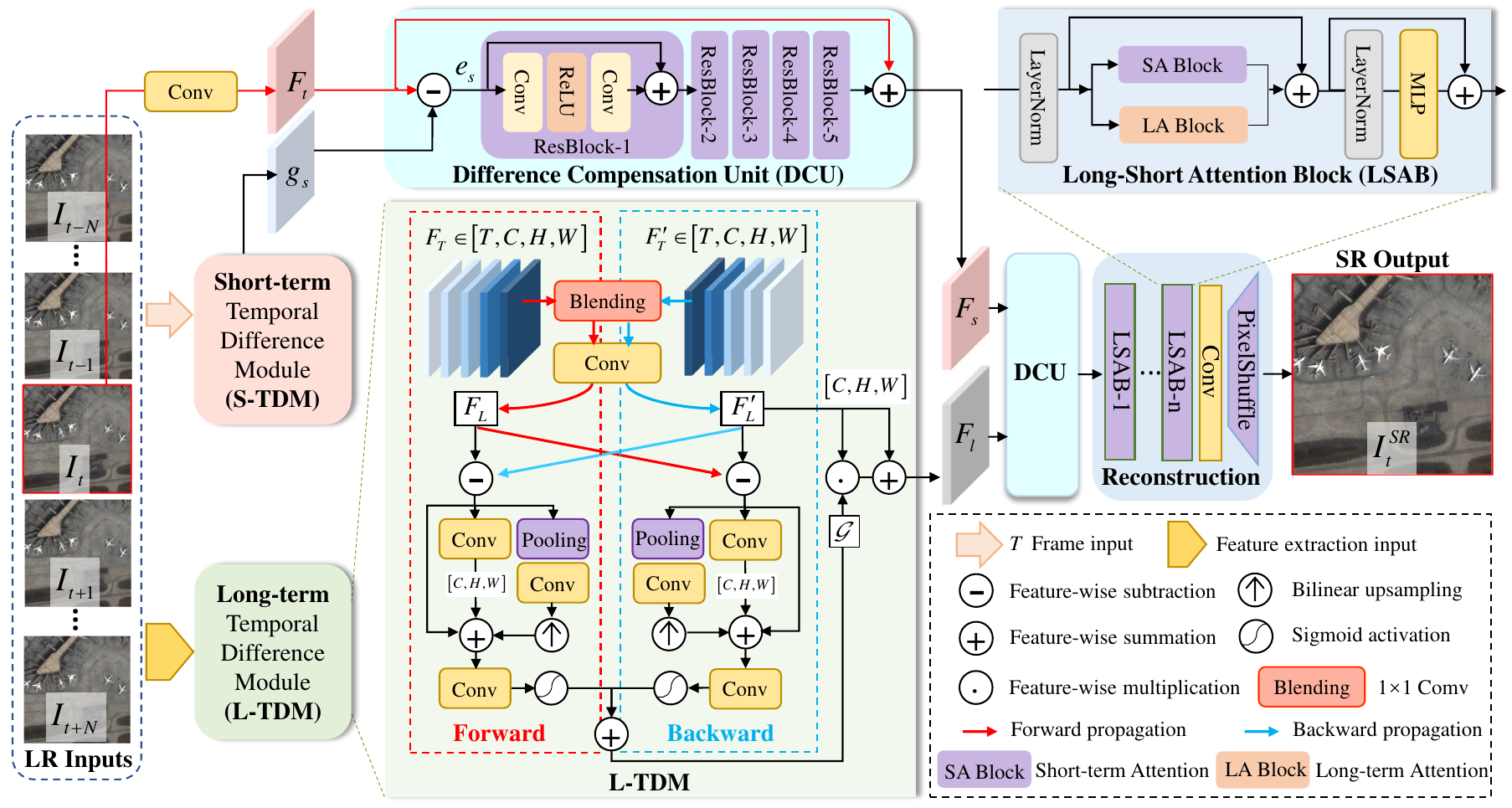}%
\captionsetup{font={scriptsize}}   
\caption{The overall structure of our proposed Local-Global Temporal Difference learning network (LGTD). It consists of four modules: (1) Short-term Temporal Difference Module (S-TDM), which is used for local temporal compensation; (2) Long-term Temporal Difference Module (L-TDM), which is proposed to realize global temporal compensation; (3) Difference Compensation Unit (DCU), which is utilized for integrating the spatial and temporal information to maintain the spatial consistency; (4) Reconstruction module, which is employed to generate the final HR target frame. Short-term Attention (SA) Block is equipped by Channel Attention, and Multi-head Self-Attention realizes Long-term (LA) Block.}
\label{network}
\end{figure*}

\section{Proposed Method}\label{meth}
\subsection{Overview}
The overall flowchart of our proposed framework is shown in Fig. \ref{network}. Given $2N+1$ LR inputs ${\rm I} = \left\{ {{I_{t - N}}, \cdots ,{I_t}, \cdots ,{I_{t + N}}} \right\}$, $I_t$ is the target frame required to be super-resolved, and the remaining frames are neighboring frames. We devise two branches to explore local-global complementary information separately by S-TDM and L-TDM. The difference compensation unit aims to realize the interaction between the temporal information obtained from TDMs and the spatial feature. The compensated feature is then sent to the reconstruction module to generate the final super-resolved target frame $I_t^{SR}$.
\par The Short-term Temporal difference Module (S-TDM), termed ${\rm{TD}}{{\rm{M}}_s}\left( \cdot \right)$, receives frame-wise inputs $\rm I$ and generates the short-term compensated results $g_i$, that is
\begin{equation}
{g_s} = {\rm{TD}}{{\rm{M}}_s}\left( {\rm I} \right),
\end{equation}

The spatial feature $F_t$ is extracted from $I_t$ by a $3\times3$ convolutional layer. $F_t$ and $g_s$ will be sent to the Difference Compensation Unit ${\rm{DCU}}\left( \cdot \right)$ to refine the compensated $g_s$ with the guide of $F_t$, thus we can maintain the spatial consistency and alleviate the misalignment in S-TDM. The refinement process can be written as follows:
\begin{equation}
{F_s} = {\rm{DCU}}\left( {{F_t},{g_s}} \right),
\end{equation}
where $F_s$ is the short-term temporal compensated feature. L-TDM received $T=2N+1$ features to explore the global temporal difference. Here, we use five residual blocks to extract $T$ features from $\rm I$. 
Take the forward branch in L-TDM as an example, $T$ features will be blended by a $1\times1$ convolution layer and another $3\times3$ convolution. In this manner, we can smooth the long-range features to get a holistic representation $F_L$ of the forward features. Similarly, we obtain the backward global feature ${F'_L}$ from $F'_T$, where $F'_T$ is the temporal reverse version of $F_T$. Thanks to L-TDM, we implicitly produce a temporal activation ${\cal G}$ for temporal compensation. The global temporal compensation result $F_l$ is determined by:
\begin{equation}
{F_l} = {F'_L} \odot {\cal G} + {F'_L},
\end{equation}
where $\odot$ is the channel-wise multiplication.
After that, the $F_l$ will also be refined by DCU with the guide of $F_s$. The final global-local temporal compensated feature ${\hat F_t}$ can be generated by the following:
\begin{equation}
{\hat F_t} = {\rm{DCU}}\left( {{F_s},{F_l}} \right).
\end{equation}

Finally, ${\hat F_t}$ will be reconstructed to ${I_t}^{SR}$ by the reconstruction module ${\rm{Reconstruction}}\left( \cdot \right)$:
\begin{equation}
I_t^{SR} = {\rm{Reconstruction}}\left( {{{\hat F}_t}} \right).
\end{equation}

\subsection{Short-term Temporal Difference Modeling}
It is observed that the motion information in short-term differences is not prominent due to the paucity of motion pixels in satellite videos. Therefore, it is inefficient to compensate for each short-term difference separately as they are extremely sparse. To address this, we choose to concatenate and enhance all the short-term differences together to explore local motion representation. This allows us to supply the target frame with a more robust representation to achieve local compensation.

Specifically, S-TDM operates feature extraction from short-term RGB difference maps between adjacent frames and propagates local motions into the target feature with lateral connections. As shown in Fig. 3., for simplicity, we set $N=2$. Firstly, we compute four temporal differences by subtracting adjacent frames. Then we extract features from these RGB differences and reduce them to low-resolution space. In this manner, we could obtain four representations $\left\{ {{d^{ - 2}},{d^{ - 1}},{d^{ + 1}},{d^{ + 2}}} \right\}$ of local motions.

As mentioned before, short-term RGB temporal differences have overwhelmingly low response regions and are prominent in motion pixels. Therefore, we argue that RGB differences are sparse signals, and it is sufficient to process them in low-resolution space. In addition, we can reduce the computational cost of further temporal difference fusion by the pooling operation. To make the target feature $f_t$ aware of short-term motion, we use a two-stage compensation strategy and supply $f_t$ with lateral connections. In the early stage, the fused temporal representation $D_s$ will be upsampled to the original size and added to the target features, which can be expressed as:
\begin{equation}
{f_1}{\rm{ = }}{f_t}{\rm{ + U}}{{\rm{p}}^ \uparrow }\left( {{\rm{Fusion}}\left( {\left[ {{d^{ - 2}},{d^{ - 1}},{d^{ + 1}},{d^{ + 2}}} \right]} \right)} \right),
\end{equation}
where ${\rm{U}}{{\rm{p}}^ \uparrow }$ denotes $\times2$ bilinear upsampling,  $\rm{Fusion}\left(\cdot \right)$ represents a $3\times3$ convolution layer, and $\left[  \cdot  \right]$ is channel concatenation. In the second stage, the fused representation $D_s$ and $f_1$ will be further extracted for deep aggregation. Finally, the short-term compensated feature $g_s$ can be written as:
\begin{equation}
{g_s} = {\rm{Re}}{{\rm{s}}_1}\left( {{f_1}} \right) + {\rm{U}}{{\rm{p}}^ \uparrow }\left( {{\rm{Re}}{{\rm{s}}_2}\left( {{D_s}} \right)} \right),
\end{equation}
where ${\rm{Re}}{{\rm{s}}_1}\left(  \cdot  \right)$ and ${\rm{Re}}{{\rm{s}}_2}\left(  \cdot  \right)$ are two residual blocks.

\begin{figure}[!t]
\centering
\includegraphics[width=3.5in]{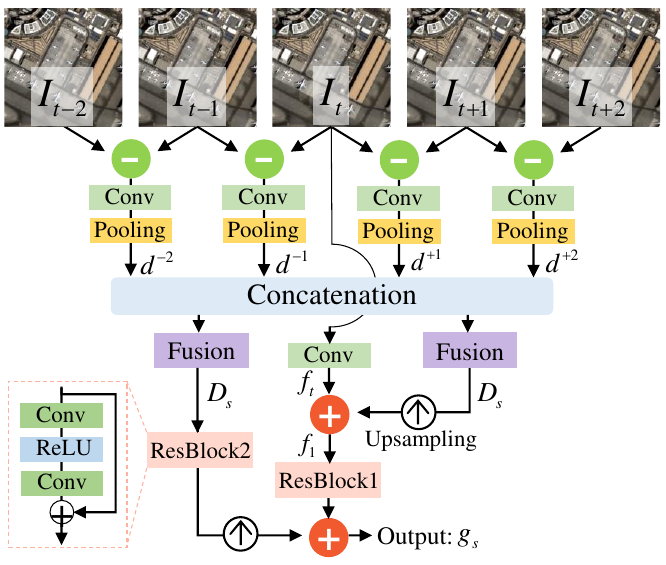}%
\captionsetup{font={scriptsize}}   
\caption{The diagram of our proposed Short-term Temporal Difference Module (S-TDM) is shown, with $N=2$ taken as an example. S-TDM performs feature extraction on stacked RGB difference maps and supplies the local motion representations into $f_t$ for local compensation.}
\label{stdm}
\end{figure}

\begin{figure}[t]
\removelatexerror 
\begin{algorithm*}[H]
\caption{Local-Global Temporal Difference Learning Algorithm.}
\label{algorithm}
\LinesNumbered 
\KwIn{LR frames ${\rm I} = \left\{ {{I_{t - N}}, \cdots ,{I_t}, \cdots ,{I_{t + N}}} \right\}$.}
\KwOut{The super-resolved target frame $I^{SR}_t$}

{\bf Initialization:} $N=5$, $S=3$, $\sigma \left(  \cdot  \right)$ is $1\times 1$ convolution, $\alpha = \beta = 0.5$.\\
\tcp{Short-term Temporal Difference}

${f_t} = {\rm{Conv}}\left( {{I_t}} \right)$;\\
\For{ $i = - N:+ N$ and $i \ne 0$}{
\eIf{$i < 0$}{${d^i} = {\rm{Pooling}}\left( {{\rm{Conv}}\left( {{I_{t + i}} - {I_{t + i + 1}}} \right)} \right)$\;}{${d^i} = {\rm{Pooling}}\left( {{\rm{Conv}}\left( {{I_{t + i}} - {I_{t + i - 1}}} \right)} \right)$\;}

}

${D_s} = Conv\left( {\left[ {{d^{ - N}}:{d^{ - N}}} \right]} \right)$;\tcp*[f]{Local motion}\\
${f_1}{\rm{ = }}{f_t}{\rm{ + U}}{{\rm{p}}^ \uparrow }\left( {{D_s}} \right)$;\tcp*[f]{One stage}\\
${g_s} = {\rm{Re}}{{\rm{s}}_1}\left( {{f_1}} \right) + {\rm{U}}{{\rm{p}}^ \uparrow }\left( {{\rm{Re}}{{\rm{s}}_2}\left( {{D_s}} \right)} \right)$;\tcp*[f]{Two stage}\\

\tcp{Coarse Alignment}
${F_T}{\rm{ = MSD}}\left( {{\rm{Res}}\left( {\rm I} \right)} \right),{F'_T} = {\rm{reverse}}\left( {{F_T}} \right)$;

\tcp{Long-term Temporal Difference}
${F_L} = {\rm{Conv}}\left( {\sigma \left( {{F_T}} \right)} \right),{F'_L} = {\rm{Conv}}\left( {\sigma \left( {{{F'}_T}} \right)} \right)$;\\
${D_f} = {F_L} - {F'_L},{D_b} = {F'_L} - {F_L}$;

$at{t_f} = {\rm{Sigmoid}}\left( {{\rm{Conv}}\left( {\sum\limits_{i = 1}^S {{H_i}\left( {{D_f}} \right)} } \right)} \right)$,\\
$at{t_b} = {\rm{Sigmoid}}\left( {{\rm{Conv}}\left( {\sum\limits_{i = 1}^S {{H_i}\left( {{D_b}} \right)} } \right)} \right)$;\\

${\cal G} = \alpha  * at{t_f} + \beta  * at{t_b}$;\tcp*[f]{Global motion}\\
${F_l} = {F'_L} \odot {\cal G} + {F'_L}$;

\tcp{Refinement}
${F_t} = {\rm{Conv}}\left( {{I_t}} \right)$;\\
${F_s} = {F_t} + {\rm{ResNet}}\left( {{F_t} - {g_s}} \right)$;\\
${\hat F_t} = {F_s} + {\rm{ResNet}}\left( {{F_s} - {F_l}} \right)$;\\

\tcp{Reconstruction}
$I_t^{SR} = {\rm{Upsample}}\left( {{\rm{Conv}}\left( {{\rm{LSA}}{{\rm{B}}^N}\left( {{{\hat F}_t}} \right)} \right)} \right)$.

\end{algorithm*}
\end{figure}

\subsection{Long-term Temporal Difference Module}
Long-term temporal differences span over a larger number of frames and can offer more pronounced motion cues. However, solely modeling the differences from long-distance frames is inadequate for exploring holistic motion patterns. Therefore, we propose to model the differences between the temporally fused forward and backward sequences, which enables us to comprehensively explore the global motion information. Since generating the global motion representation is more complicated, we adopt a holistic activation approach to modulate the temporal feature, which enables robust global compensation.

To handle the large motion while mitigating the misalignment within long-distance frames, we employ a shadow alignment operation through our previous Multi-Scale DConv alignment module \cite{9530280}. Thus, the coarsely aligned feature $F_T$ and $F'_T$ could be compressed to smooth the large variance. Furthermore, the temporal difference is calculated between $F_T$ and $F'_T$, \emph{i.e.}, ${D_f} = {F_L} - {F'_L}$ and ${D_b} = {F'_L} - {F_L}$. This cross-difference operation encourages forward and backward propagation to utilize complementary information in different segments.

During the succeeding propagation, a multi-scale design is employed to better preserve the multi-scale information in satellite video. In particular, three operations at different scales will be applied to the global temporal differences $D_f$ and $D_b$: (1) Identification connection to stabilize the gradient; (2) Deep feature extraction on the original scale by a $3\times3$ convolution; (3) Feature propagation in a small-to-large scale, achieved by a pooling layer, a $3\times3$ convolution, and a bilinear upsampling. The outputs of three operations will be aggregated and sent to a convolution layer and sigmoid function for activation:
\begin{gather}
at{t_f} = {\rm{Sigmoid}}\left( {{\rm{Conv}}\left( {\sum\limits_{i = 1}^3 {{H_i}\left( {{D_f}} \right)} } \right)} \right),\\
at{t_b} = {\rm{Sigmoid}}\left( {{\rm{Conv}}\left( {\sum\limits_{i = 1}^3 {{H_i}\left( {{D_b}} \right)} } \right)} \right),
\end{gather}
where $H_i$ represents the $i$-th operations mentioned above. $at{t_f}$ and $at{t_b}$ are the activated maps in froward and backward branches. We composed $at{t_f}$ and $at{t_b}$ to obtain a final activation ${\cal G}$, that is:
\begin{equation}
{\cal G} = \alpha  * at{t_f} + \beta  * at{t_b},
\end{equation}
where $ \alpha$ and $\beta$ are two balance coefficient. Here we set $\alpha = \beta = 0.5$. In the end, the long-term temporal compensated feature $F_l$ is generated by feature modulation:
\begin{equation}
{F_l} = {F'_L} \odot {\cal G} + {F'_L},
\end{equation}
where $\odot$ is channel-wise multiplication.

\subsection{Difference Compensation Unit}
Benefiting from our S-TDM and L-TDM, we can systematically explore global and local temporal information. However, misalignment can still occur due to the inherent limitations of temporal difference learning. Towards this end, the compensated feature $F_s$ and $F_l$ need to be further refined to preserve valuable information to spatial feature $F_t$ of the target frame. With the help of DCU, the interference in temporal difference learning could be eliminated, which is beneficial for reducing artifacts caused by misalignment.
\par As shown in Fig. \ref{network}, by predicting the difference feature between spatial feature $F_t$ and short-term temporal compensated feature $g_s$, the residual feature $e_s$ will be further enhanced and supplied to $F_t$. In this manner, the S-TDM will be forced to focus on the valuable compensation information of $F_t$. The difference compensation between $F_t$ and $g_s$ can be summarized as:
\begin{equation}
{F_s} = {F_t} + {\rm{ResNet}}\left( {{F_t} - {g_s}} \right),
\end{equation}
Similarly, $F_s$ will be compensated by $F_l$ for the final refinement:
\begin{equation}
{\hat F_t} = {F_s} + {\rm{ResNet}}\left( {{F_s} - {F_l}} \right),
\end{equation}

\begin{table*}[h]
  \centering
\captionsetup{font={scriptsize}}   
  \caption{Quantitative comparisons on Jilin-T. The PSNR/SSIMs are calculated on the luminance channel (Y). The FLOPs are computed on LR inputs with the size of $160\times 160$. Note that BasicVSR and BasicVSR++ are recurrent propagation methods. Here their propagate length is set to 15. The best and second performances are highlighted in \textcolor{red}{red} and \textcolor{blue}{blue}, respectively.}

\renewcommand\arraystretch{1.2}  
\setlength{\tabcolsep}{1.5mm}{  
    \begin{tabular}{c|ccc|cccccc}
\hline
    Method & Frames & \#Param. (M) & FLOPs (G) & Scene-1 & Scene-2 & Scene-3 & Scene-4 & Scene-5 & Average \\
\hline
    Bicubic & - & -     &-       & 31.05/0.9097 & 28.57/0.8630 & 30.62/0.9009 & 33.72/0.9391 & 32.47/0.9246 & 31.29/0.9075 \\
    TDAN \cite{tian2020tdan}  & 5 & 1.97  &-       & 35.32/0.9577 & 31.72/0.9255 & 34.08/0.9477 & 38.43/0.9727 & 36.37/0.9626 & 35.18/0.9532 \\
    DUF-52L \cite{jo2018deep} &7 & 6.8   &736.6       & 35.83/0.9604 & 32.29/0.9326 & 34.57/0.9525 & 39.15/0.9761 & 36.88/0.9659 & 35.74/0.9575 \\
    RBPN \cite{haris2019recurrent} &7 & 12.8  &3785.3       & 35.73/0.9595 & 32.10/0.9307 & 34.65/0.9534 & 39.26/0.9767 & 36.90/0.9660 & 35.73/0.9573 \\
    EDVR-L \cite{wang2019edvr}  &5 & 20.7  &897.8       & 36.05/0.9620 & 32.45/0.9353 & 34.87/0.9554 & 39.37/0.9769 & 36.82/0.9660 & 35.91/0.9591 \\
    SOF-VSR \cite{wang2020deep} &3 & 1.06   &127.3       & 35.89/0.9610 & 32.26/0.9328 & 34.70/0.9537 & 39.15/0.9760 & 36.88/0.9660 & 35.78/0.9579 \\
    MSDTGP \cite{9530280} &5 & 14.1  &1579.8       & 36.13/0.9631 & 32.42/0.9350 & 34.81/0.9551 & 39.46/0.9773 & 37.10/0.9675 & 35.98/0.9600 \\
    MANA \cite{yu2022memory}  &7 & 22.2  &633.5       & 35.94/0.9616 & 32.43/0.9347 & 34.78/0.9544 & 39.34/0.9768 & 37.03/0.9670 & 35.90/0.9589 \\
    BasicVSR \cite{chan2021basicvsr} &15 & 6.3   &163.7       & \textcolor{blue}{36.14}/0.9631 & \textcolor{red}{32.60/0.9372} & 34.87/0.9551 & 39.40/0.9770 & \textcolor{blue}{37.19}/0.9677 & 36.04/0.9601 \\
BasicVSR++ \cite{chan2022basicvsr++} & 15 & 7.0 & 2719.4 & 36.12/\textcolor{blue}{0.9633} & \textcolor{blue}{32.59}/0.9367 & \textcolor{blue}{34.88/0.9556} & \textcolor{red}{39.48/0.9777} &  37.17/\textcolor{red}{0.9684} & \textcolor{blue}{36.05/0.9603} \\
    \textbf{LGTD (Ours)}  &5 & 20.7  &647.8       & \textcolor{red}{36.21/0.9636} & 32.58/\textcolor{blue}{0.9369} & \textcolor{red}{34.93/0.9559} & \textcolor{blue}{39.47/0.9773} & \textcolor{red}{37.22}/\textcolor{blue}{0.9682} & \textcolor{red}{36.08/0.9604} \\
\hline
    \end{tabular}%
  \label{jilin}%
}
\end{table*}%

\begin{table*}[h]
  \centering
\captionsetup{font={scriptsize}}   
  \caption{Quantitative comparisons on Carbonite-2 and UrtheCast. The best and second performances are highlighted in \textcolor{red}{red} and \textcolor{blue}{blue}, respectively.}
\renewcommand\arraystretch{1.2}  
    \begin{tabular}{c|c|c|ccccc}
\hline
            Data Source & Method & Publication & Scene-6 & Scene-7 & Scene-8 & Scene-9 & Average \\
\hline
    \multirow{9}[0]{*}{Carbonite-2} & Bicubic & -&39.23/0.9443 & 36.68/0.9201 & 32.25/0.8754 & 35.59/0.9271 & 35.94/0.9167 \\
          & DUF-52L \cite{jo2018deep} & \emph{CVPR'18}&41.47/0.9623 & 39.14/0.9471 & 34.74/0.9208 & 38.43/0.9528 & 38.45/0.9458 \\
          & SOF-VSR \cite{wang2020deep} & \emph{TIP'20}&41.32/0.9611 & 38.96/0.9456 & 34.55/0.9178 & 38.20/0.9514 & 38.26/0.9440 \\
          & RBPN \cite{haris2019recurrent}  & \emph{CVPR'19}&41.44/0.9624 & 39.04/0.9465 & 34.92/0.9237 & 38.58/0.9537 & 38.50/0.9466 \\
          & EDVR-L \cite{wang2019edvr} & \emph{CVPRW'19}&41.59/0.9632 & 39.21/0.9477 & 35.09/0.9251 & 38.66/0.9545 & 38.64/0.9476 \\
          & MSDTGP \cite{9530280} & \emph{TGRS'22}&41.56/0.9630 & 39.00/0.9461 & 34.85/0.9225 & 38.48/0.9532 & 38.47/0.9462 \\
          & MANA \cite{yu2022memory}  & \emph{CVPR'22}&\textcolor{blue}{41.66}/0.9634 & 39.12/0.9470 & 34.89/0.9224 & 38.66/0.9541 & 38.58/0.9468 \\
          & BasicVSR \cite{chan2021basicvsr} & \emph{CVPR'21}&41.64/\textcolor{blue}{0.9634} & 39.21/0.9478 & 35.10/0.9266 & 38.83/\textcolor{blue}{0.9550} & 38.70/0.9482 \\
          & BasicVSR++ \cite{chan2022basicvsr++} & \emph{CVPR'22}&41.66/0.9630
 & \textcolor{blue}{39.33/0.9505} & \textcolor{blue}{35.15/0.9290} & \textcolor{blue}{38.86}/0.9549 & \textcolor{blue}{38.75/0.9493} \\
          & \textbf{LGTD (Ours)} & -&\textcolor{red}{41.73/0.9643} & \textcolor{red}{39.52/0.9516} & \textcolor{red}{35.27/0.9302} & \textcolor{red}{38.90/0.9557} & \textcolor{red}{38.86/0.9504} \\
\hline
    Data Source & Method & Publication & Scene-10 & Scene-11 & Scene-12 & Scene-13 & Average \\
\hline
    \multirow{9}[1]{*}{UrtheCast} & Bicubic & - & 43.86/0.9917 & 25.32/0.7242 & 33.43/0.9130 & 45.20/0.9939 & 36.95/0.9057 \\
          & DUF-52L \cite{jo2018deep} & \emph{CVPR'18}& 46.12/0.9939 & 26.66/0.7964 & 35.51/0.9407 & 47.57/0.9958 & 38.97/0.9317 \\
          & SOF-VSR \cite{wang2020deep} & \emph{TIP'20}& 45.68/0.9934 & 26.67/0.7934 & 35.46/0.9397 & 46.70/0.9952 & 38.63/0.9304 \\
          & RBPN \cite{haris2019recurrent}  & \emph{CVPR'19} & 45.84/0.9936 & 26.66/0.7965 & 35.43/0.9404 & 47.14/0.9954 & 38.77/0.9315 \\
          & EDVR-L \cite{wang2019edvr} & \emph{CVPRW'19} &46.10/\textcolor{red}{0.9940} & 26.89/0.8036 & \textcolor{red}{35.65}/\textcolor{blue}{0.9417} & 47.84/\textcolor{blue}{0.9960} & \textcolor{blue}{39.12}/0.9338 \\
          & MSDTGP \cite{9530280} & \emph{TGRS'22} &46.08/0.9939 & 26.75/0.8010 & 35.46/0.9409 & 47.39/0.9957 & 38.92/0.9329 \\
          & MANA \cite{yu2022memory}  & \emph{CVPR'22} &46.02/0.9938 & 26.89/0.8040 & 35.51/0.9412 & 47.56/0.9957 & 39.01/0.9337 \\
          & BasicVSR \cite{chan2021basicvsr} & \emph{CVPR'21}&\textcolor{blue}{46.13}/\textcolor{blue}{0.9940} & 26.97/0.8059 & 35.54/0.9416 & 47.83/0.9959 & 39.12/0.9343 \\
          & BasicVSR++ \cite{chan2022basicvsr++} & \emph{CVPR'22}&46.09/0.9938
 & \textcolor{blue}{27.09/0.8069} &  \textcolor{blue}{35.58}/\textcolor{red}{0.9418} & \textcolor{blue}{47.86}/0.9957 & 39.12/\textcolor{blue}{0.9346} \\
          & \textbf{LGTD (Ours)} & -& \textcolor{red}{46.20}/0.9939 & \textcolor{red}{27.04/0.8084} &35.54/0.9413 & \textcolor{red}{48.09/0.9960} & \textcolor{red}{39.22/0.9349} \\
\hline
    \end{tabular}%
  \label{cbu}%
\end{table*}%

\subsection{Reconstruction}
After short-term and long-term temporal information compensation, the spatial-temporal information is entangled. Therefore, we can not reconstruct the compensated feature with a naive design, such as using residual blocks. Nowadays, the attention mechanism has been widely used in super-resolution reconstruction, with the success of Channel Attention (CA) in RCAN \cite{zhang2018image} and Multi-head Self-Attention (MSA) in SwinIR \cite{liang2021swinir}. Previous work \cite{chen2022activating} has demonstrated that CA and MSA are experts in learning short-term and long-term dependencies, respectively. Here, we comfortably adopt the widely used CA and MSA to reconstruct the entangled local-global information in a hybrid design. In Fig. \ref{network}, the Short-term Attention block (SA Block) is employed by the CA module in RCAN, and the Long-term Attention block (LA Block) is equipped with MSA in SwinIR. Finally, five Long-Short Attention Blocks (LSAB) and a Pixel-Shuffle layer are stacked to restore the super-resolved target frame:
\begin{equation}
I_t^{SR} = {\rm{Upsample}}\left( {{\rm{Conv}}\left( {{\rm{LSA}}{{\rm{B}}^N}\left( {{{\hat F}_t}} \right)} \right)} \right).
\end{equation}

\begin{figure*}[h]
\centering
\includegraphics[width=7in]{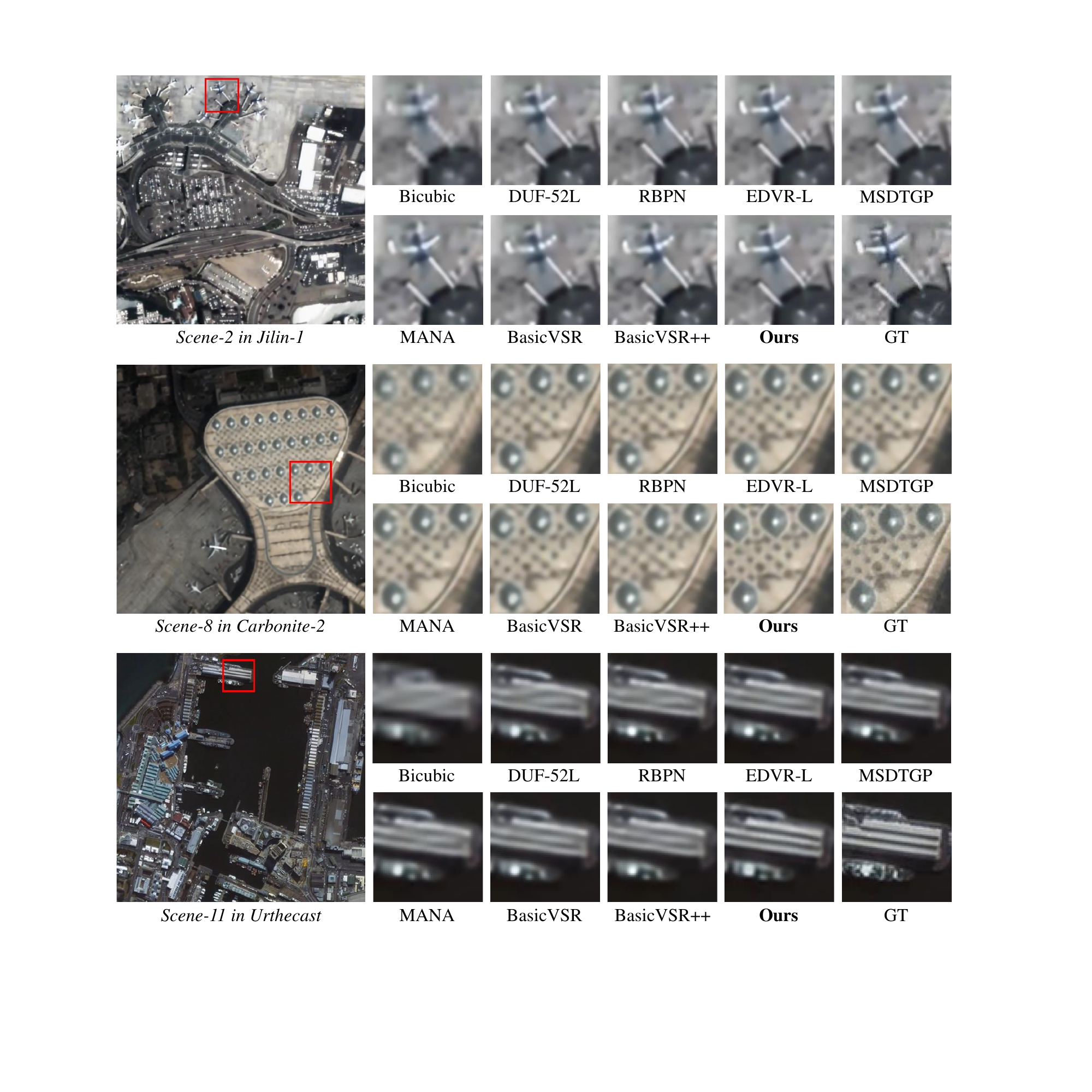}%
\captionsetup{font={scriptsize}}   
\caption{Qualitative comparisons on scene-2 of Jilin-1, scene-8 of Carbonite-2, and scene-11 from UrtheCast. Zoom in for better visualization.}
\label{fig-jilin}
\end{figure*}

\begin{figure*}[h]
\centering
\includegraphics[width=7in]{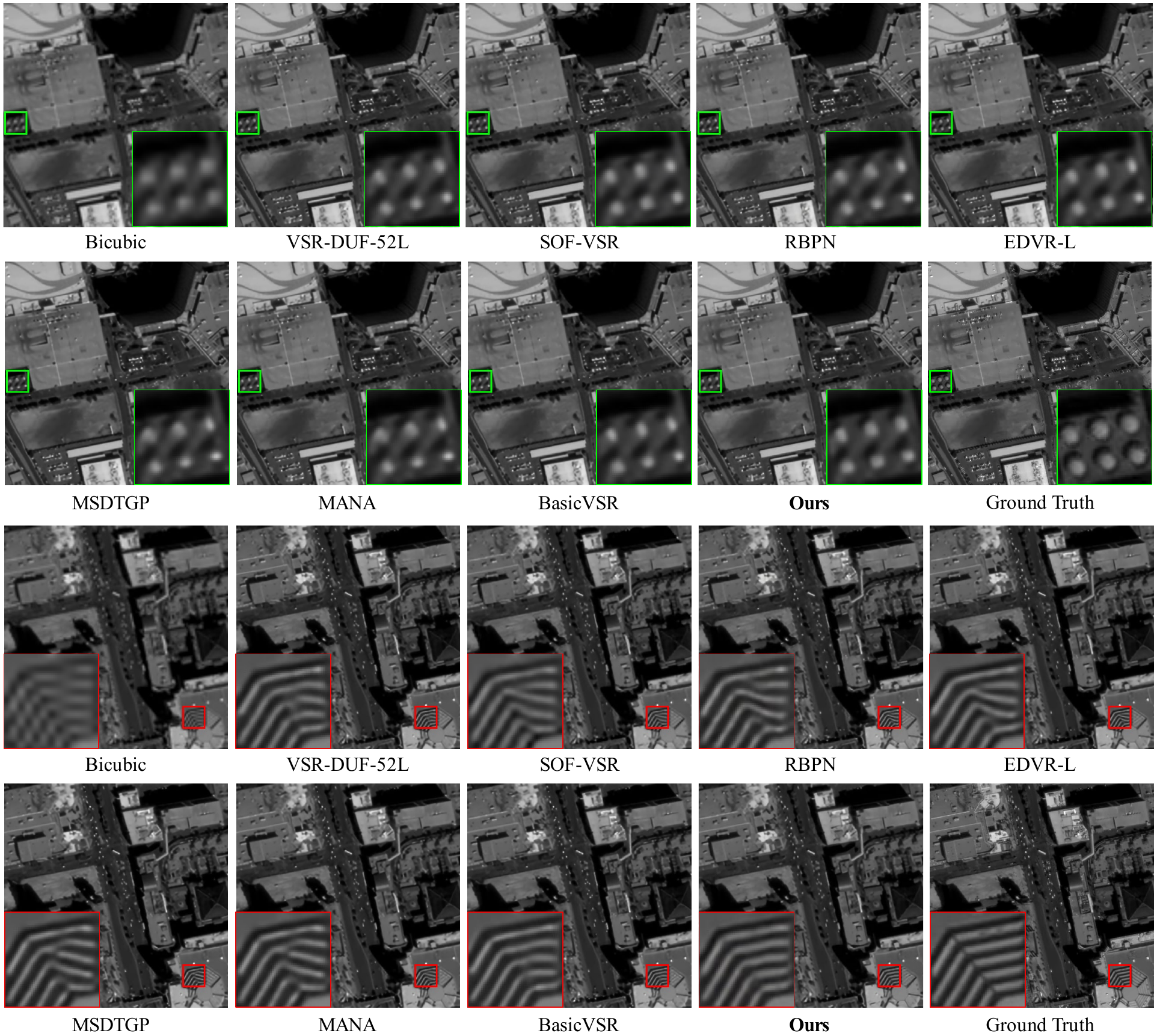}%
\captionsetup{font={scriptsize}}   
\caption{Qualitative comparisons on scene-14 and scene-15 of SkySat-1. Zoom in for better visualization.}
\label{fig-sky}
\end{figure*}

\begin{table*}[h]
  \centering
  \captionsetup{font={scriptsize}}   
  \caption{Quantitative comparisons on SkySat-1 and Zhuhai-1. The best and second performances are highlighted in \textcolor{red}{red} and \textcolor{blue}{blue}, respectively.}
\renewcommand\arraystretch{1.2}  
\setlength{\tabcolsep}{1.0mm}{  
    \begin{tabular}{c|c|cccccccc}
\hline
    Data Source & Scene & DUF-52L \cite{jo2018deep}  & RBPN \cite{haris2019recurrent}  & \multicolumn{1}{c}{MSDTGP \cite{9530280}} & EDVR-L \cite{wang2019edvr} & MANA \cite{yu2022memory}  & BasicVSR \cite{chan2021basicvsr} & BasicVSR++ \cite{chan2022basicvsr++}& \textbf{LGTD (Ours)} \\
\hline
    \multicolumn{1}{c|}{\multirow{2}[0]{*}{SkySat-1}} & Scene-14     & 33.77/0.9302  & 33.79/0.9301 & \multicolumn{1}{c}{33.79/0.9317} & 33.98/0.9339 & 33.94/0.9331 & \textcolor{blue}{34.00/0.9343} & 33.98/0.9340& \textcolor{red}{34.03/0.9360} \\
          & Scene-15     & 33.34/0.9136  & 33.24/0.9123 & \multicolumn{1}{c}{33.23/0.9141} & 33.42/\textcolor{blue}{0.9157} & 33.27/0.9138 & 33.38/0.9144 & \textcolor{blue}{33.47}/0.9150&\textcolor{red}{33.66/0.9198} \\
\hline
    Zhuhai-1 & Scene-16     & 32.85/0.9002 & 32.71/0.8976 & \multicolumn{1}{c}{32.83/0.9004} & 32.85/0.9013 & 32.80/0.9003 & 32.80/0.9006 &\textcolor{blue}{32.86/0.9016}& \textcolor{red}{33.04/0.9048} \\
\hline
    \multicolumn{2}{c|}{Average} & 33.32/0.9147 & 33.25/0.9133 & 33.28/0.9154 & 33.42/\textcolor{blue}{0.9170} & 33.34/0.9157 & 33.39/0.9164 & \textcolor{blue}{33.44}/0.9169& \textcolor{red}{33.58/0.9202} \\
\hline
    \end{tabular}%
}
  \label{sz}%
\end{table*}%

\begin{figure*}[h]
\centering
\includegraphics[width=7in]{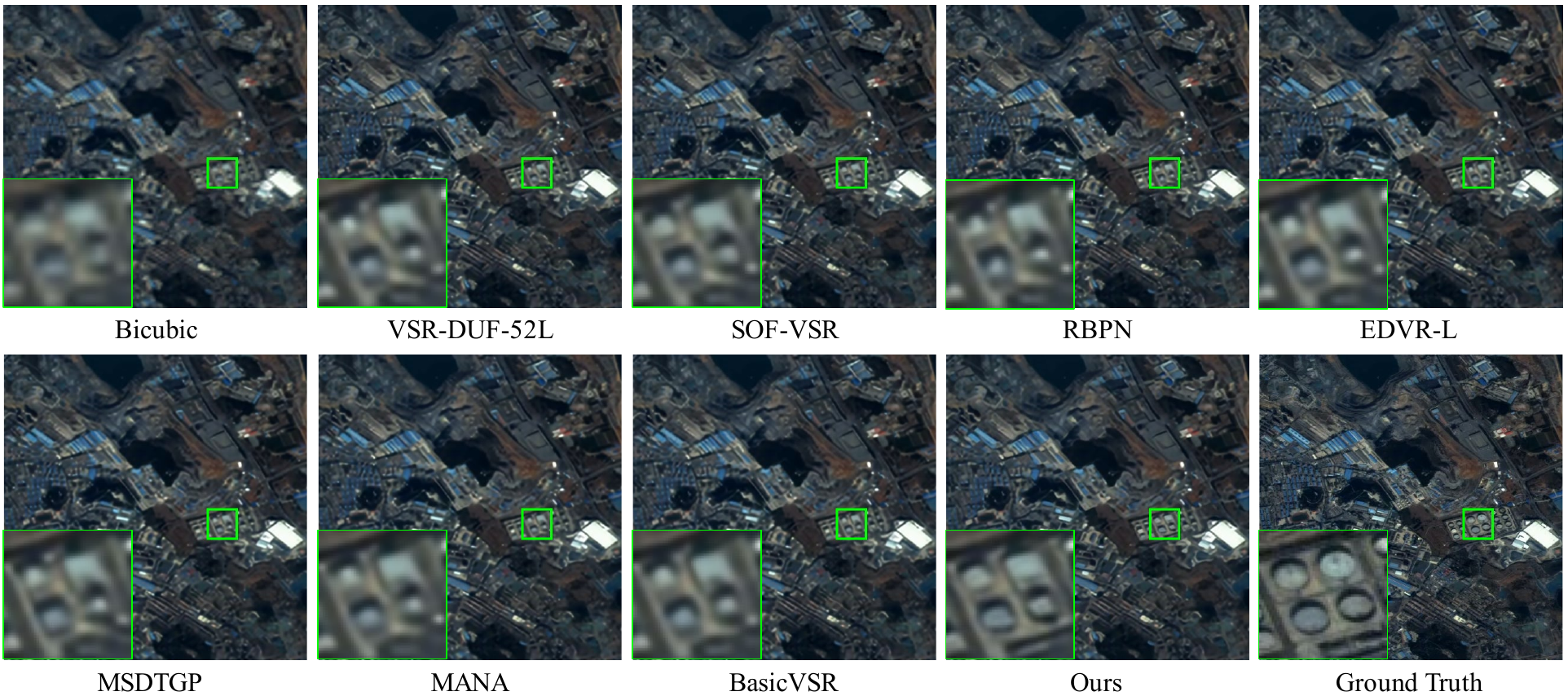}%
\captionsetup{font={scriptsize}}   
\caption{Qualitative comparisons on scene-16 of Zhuhai-1. Zoom in for better visualization.}
\label{fig-zhuhai}
\end{figure*}

\begin{figure}[h]
\centering
\includegraphics[width=3.5in]{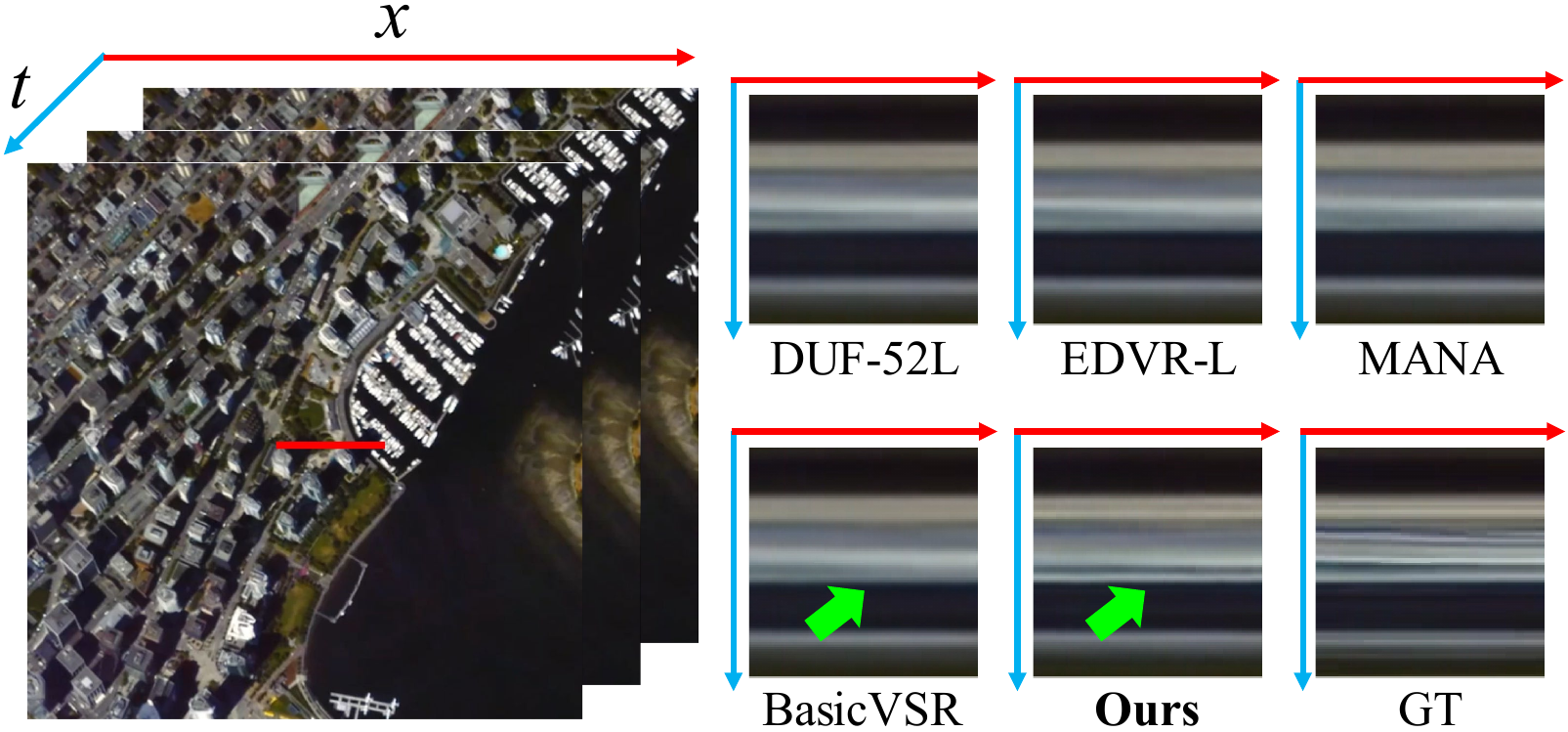}%
\captionsetup{font={scriptsize}}   
\caption{Qualitative comparisons of temporal profile. The temporal profile is generated by recording a pixel line (red light) with stacking among frames.}
\label{profile}
\end{figure}

\begin{figure}[h]
\centering
\includegraphics[width=3.5in]{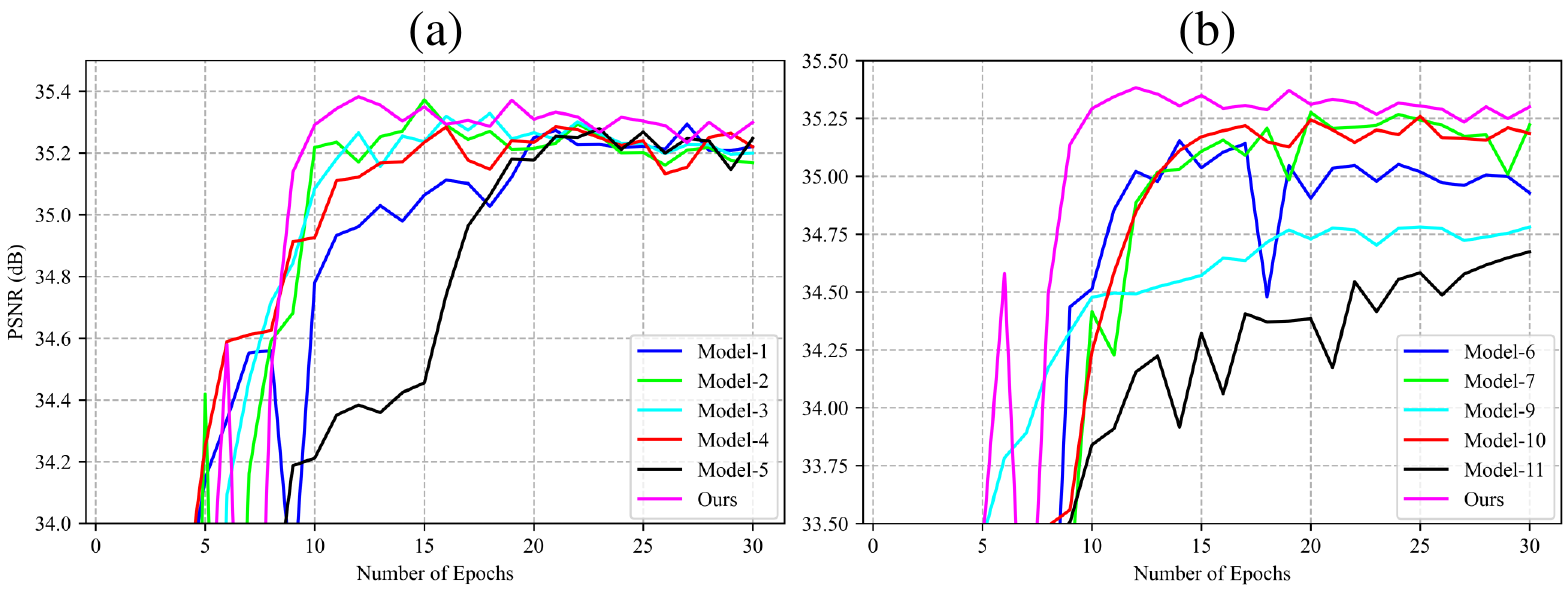}%
\captionsetup{font={scriptsize}}   
\caption{The training process of different models. Model-1 and Model-2 are used to validate the effectiveness of S-TDM and L-TDM. Model-3, Model-4, and Model-5 are designed to prove the effectiveness of calculating differences. Model-6 and Model-7 are used to demonstrate the effectiveness of bidirectional propagation in L-TDM. Finally, Model-9, Model-10, and Model-11 are established to indicate the effectiveness of hybrid attention.}
\label{training}
\end{figure}

\begin{figure}[h]
\centering
\includegraphics[width=3.5in]{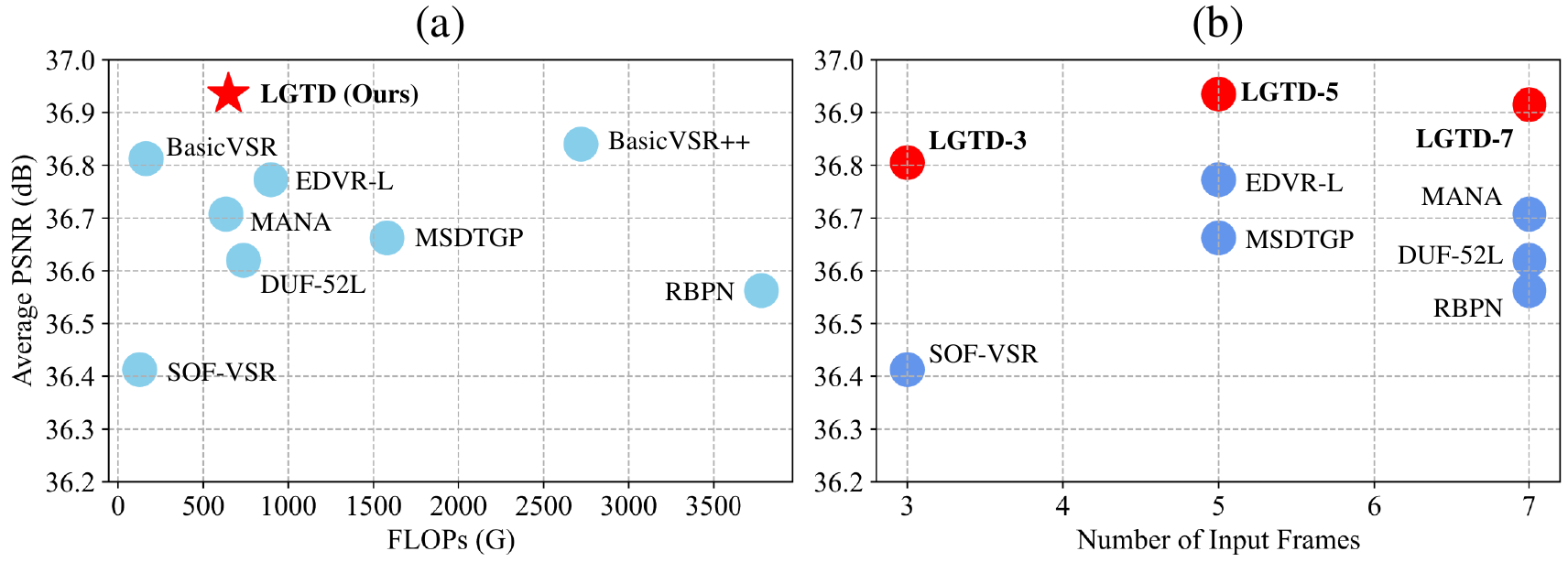}%
\captionsetup{font={scriptsize}}   
\caption{(a) displays the relationship between FLOPs and PSNR of each model. (b) shows the relationship between input frame numbers and PSNR. Our LGTD achieves the best performance with acceptable FLOPs.}
\label{number}
\end{figure}

\section{Experiment and Discussion}\label{exp}
\subsection{Satellite Video Datasource}
To comprehensively evaluate our LGTD, we collected extensive satellite video clips from five mainstream video satellites, including Jilin-1, Carbonite-2, UrtheCast, SkySat-1, and Zhuhai-1. Following our previous work \cite{9530280}, we cropped 189 clips of $640\times 640$ from Jilin-1 to build the training set. Five scenes are randomly cropped from two Jilin-1 videos to build the test set Jilin-T. Besides, we additionally crop four scenes from Carbonite-2 and UrtheCast, respectively. As for SkySat-1 and Zhuhai-1, two and one scenes are selected for further testing. Eventually, we have 189 video clips used for model training and 16 scenes from five satellites to evaluate model performance. The training and test sets can be found at \url{https://github.com/XY-boy/MSDTGP}

\subsection{Implementation Details}
Denoting the size of $I^{SR}_t$ as $Hr\times Wr\times c$, where $c=3$ means the RGB channels, $H$ and $W$ represent the height and width of LR input, and $r$ is the scale factor. In this paper, we only focus on $r=4$. The number of input LR frames is set to 5. During model training, we sample 4 LR video patches with size $64\times64$ in each mini-batch. Data augmentation is realized on LR inputs by random flip and rotation. The initial learning rate is set to $1\times10^{-4}$ and decays to half the previous one when it reaches every ten epochs. To minimize the ${{\cal L}_1} = {\left\| {I_t^{SR} - I_t^{GT}} \right\|_1}$ distance between $I^{SR}_t$ and ground-truth frame $I^{GT}_t$, the Adam optimization with ${\beta _1} = 0.9$ and ${\beta _2} = 0.999$ was used. We trained our model on a single NVIDIA RTX 3090 for 50 epochs, and it took us nearly 40 hours for model training.

\subsection{Metrics}
\subsubsection{Peak-Signal-to-Noise Ratio (PSNR)} Here, we adopt PSNR as the objective metric for image quality evaluation. Given a reference frame $I_t^{GT}$ and a test frame $I_t^{SR}$, both of size $H\times W$, the PSNR between them is defined by:
\begin{equation}
PSNR\left( {I_t^{GT},I_t^{SR}} \right) = 10{\log _{10}}\left( {{{255}^2}/MSE\left( {I_t^{GT},I_t^{SR}} \right)} \right),
\end{equation}
where
\begin{equation}
MSE\left( {I_t^{GT},I_t^{SR}} \right) = \frac{1}{{HW}}\sum\limits_{i = 1}^H {\sum\limits_{j = 1}^W {{{\left( {I_t^{GT},I_t^{SR}} \right)}^2}} }.
\end{equation}
A high value of the PSNR implies low numerical differences between $I_t^{GT}$ and $I_t^{SR}$, which means the higher the PNSR, the better the reconstruction quality.
\subsubsection{Structural Similarity Index Measure (SSIM)}SSIM is used to measure the similarity between $I_t^{GT}$ and $I_t^{SR}$. Different from PSNR which uses error summation, SSIM is determined by the combination of three factors that are luminance $l\left(  \cdot  \right)$, contrast $c\left(  \cdot  \right)$, and structure $s\left(  \cdot  \right)$ comparison function. The SSIM is calculated by:
\begin{equation}
SSIM\left( {I_t^{GT},I_t^{SR}} \right) = l\left( {I_t^{GT},I_t^{SR}} \right)c\left( {I_t^{GT},I_t^{SR}} \right)s\left( {I_t^{GT},I_t^{SR}} \right).
\end{equation}
More details on each comparison function can be found in \cite{wang2004image}. The values of the SSIM are in $\left[ {0,1} \right]$. A value close to 1 implies a high correlation between $I_t^{GT}$ and $I_t^{SR}$.

\subsection{Comparison With State-of-the-Arts}
We compared the proposed LGTD with state-of-the-art approaches, including TDAN \cite{tian2020tdan}, DUF-52L \cite{jo2018deep}, RBPN \cite{haris2019recurrent}, EDVR-L \cite{wang2019edvr}, SOF-VSR \cite{wang2020deep}, MSDTGP \cite{9530280}, MANA \cite{yu2022memory}, BasicVSR \cite{chan2021basicvsr}, and BasicVSR++ \cite{chan2022basicvsr++}. For a fair comparison, we retrained these methods on the Jilin-1 training set following their official implementation. The PSNR and SSIM are used as objective metrics to evaluate the fidelity of the restored results. Note that the PSNR is calculated on the luminance (Y) channel \cite{ jk2}, and we crop 8 pixels on the boundary as \cite{jo2018deep}.

\subsubsection{Quantitative Evaluation}
As listed in Table. \ref{jilin}, \ref{cbu}, \ref{sz}, our LGTD surpasses optical flow-based and DConv-based methods in terms of quantitative evaluation on various satellite video datasets without relying on optical flow estimation. This highlights the potential of our TDMs in overcoming the limitations of optical flow and improving the performance of temporal compensation in satellite videos. In particular, our LGTD outperforms the flow-based method SOF-VSR 0.3dB and the kernel-based method EDVR-L 0.17dB on the Jilin-T in Table.\ref{jilin}. In comparison to BasicVSR and BasicVSR++, which employ recurrent propagation for alignment, the proposed LGTD method continues to yield favorable results while utilizing a reduced number of input frames (5 frames compared to 15 frames). These results prove that our S-TDM and L-TDM can explore sufficient temporal information in a local window.

In Table. \ref{cbu}, we found that EDVR-L and BasicVSR yield approaching performance. Although the recurrent propagation-based method enlarges the temporal receptive field compared with the window-slide approach, it still can not significantly improve the PSNR while consuming more frames as input. Our LGTD outperforms BasicVSR++ by 0.11dB in the Carbonite-2 dataset and by 0.1dB in the UrtheCast dataset. This highlights the effectiveness of the proposed L-TDM in effectively exploring long-range dependencies compared to the recurrent structure. Furthermore, in Table \ref{sz}, EDVR-L achieves the second-best performance on SkySat-1 and Zhuhai-1 datasets, suggesting that the DConv-based alignment method tends to exhibit better generalization capabilities in remote sensing scenarios compared to recurrent propagation techniques. This observation can be attributed to the fact that recurrent propagation may accumulate misalignments as the number of input frames increases and the scenes become more diverse. Benefiting from the difference compensation unit, we can mitigate the misalignment by enhancing the interaction between spatial and temporal information. This, in turn, preserves spatial consistency and yields optimal reconstruction performance.

\subsubsection{Qualitative Results}
The visual comparison results are shown in Fig. \ref{fig-jilin}, \ref{fig-sky}, and \ref{fig-zhuhai}. In Fig. \ref{fig-jilin}, LGTD can restore sharper and more reliable details. Specifically, in scene-2 of Jilin-1, UDF-52L, RBPN, and SOF-VSR produce blurs on the wing of the plane. It suggests that both 3D convolution and optical flow are inadequate in providing fine-grained temporal information. In scene-11 of UrtheCast, DUF-52L and RBPN predict distorted lines. EDVR-L, MSDTGP, BasicVSR, and BasicVSR++, while relatively accurate, still appear less distinct and less clear compared to our LGTD. In scene-8 of Carbonite-2, LGTD succeeds in maintaining the accurate shape and edges of the elliptical marks on the building. As discussed earlier, short-term temporal difference maps have high responses along object boundaries, while long-term difference maps focus on the edge and shape information. The visual comparison reveals that the proposed S-TDM and L-TDM could exploit valuable high-frequency information and compensate them to the target frame. Furthermore, qualitative experiments visually reveal artifacts caused by misalignment. Both optical flow and DCN-based approaches tend to produce unrealistic distortions, while our DCU aids in preserving spatial consistency.

In Fig. \ref{fig-sky}, LGTD is capable of restoring circular ground objects and retains line information for buildings. All the methods, except BasicVSR and our LGTD, recover contorted results with severe artifacts. These results highlight the ability of TDMs to maintain temporal consistency from neighboring frames and mitigate the undesired misalignment.

In Fig. \ref{fig-zhuhai}, we zoomed in and displayed the details of a scene from Zhuhai-1. LGTD delivers comparatively realistic visual performance, whereas other approaches struggle to recover high-frequency information. To better evaluate the coherence and consistency of the restored video over long-time series dynamics, we recorded a line of pixels and stacked them on the timeline. The resulting temporal profile is shown in Fig. \ref{profile}. Our LGTD generates precise long-term temporal details, highlighting the effectiveness of our L-TDM in global compensation.

\subsection{Ablation Studies}
In this section, we conduct extensive experiments to demonstrate the effectiveness of temporal difference learning. Note that the PSNR is calculated in scene-2 of the Jinlin-1 dataset.
\subsubsection{The Number of Input Frames and Model Efficiency}We investigate the impact of varying input frame numbers for each model in Figure \ref{number}(b). Moreover, the correlation between FLOPs and PSNR for each model is shown in Fig. \ref{number}(a). Here, the PSNR values are averaged over 16 scenes within five satellite videos. LGTD-$n$ denotes the utilization of $n$ consecutive frames as input. From Fig.\ref{number}(a), we observed that LGTD strikes a favorable balance between performance and computational complexity. The results in Fig.\ref{number}(b) indicate that our LGTD consistently attains the best performance across all input frame conditions. As a result, we select five frames as input for our final model, as LGTD-5 exhibits the highest performance.

\subsubsection{Effectiveness of S-TDM and L-TDM}
To assess their contributions, we conduct two comparisons by excluding the L-TDM (Model-1) and S-TDM (Model-2). As reported in Table. \ref{abla-1}, the PSNR values show a significant reduction when we do not perform temporal difference modeling. This observation demonstrates our TDM can provide valuable local and global dependencies for temporal compensation.
\begin{table}[!t]
  \centering
  \captionsetup{font={scriptsize}}   
  \caption{Ablation experiment on the effectiveness of S-TDM and L-TDM. Results are calculated on scene 2 of Jilin-T in the Y channel.}
\renewcommand\arraystretch{1.2}  
    \begin{tabular}{c|cc|ccc}
\hline
    Models & S-TMD & L-TMD & FLOPs & \#Param. & PSNR (dB) \\
\hline
    Model-1 & \checkmark     &       & 481.02 & 18.72 & 35.28 \\
    Model-2 &       & \checkmark     & 633.83 & 19.69 & 35.34 \\
    \textbf{Ours}  & \checkmark     & \checkmark     & 647.80 & 20.73 & \textbf{35.38} \\
\hline
    \end{tabular}%
  \label{abla-1}%
\end{table}%
\begin{table}[!t]
  \centering
  \captionsetup{font={scriptsize}}   
  \caption{Ablation experiment on the effectiveness of difference operation. Results are calculated on scene 2 of Jilin-T in the Y channel.}
\renewcommand\arraystretch{1.2}  
    \begin{tabular}{c|cc|ccc}
\hline
    Models & S-TDM & L-TDM & FLOPs & \#Param. & PSNR (dB) \\
\hline
    Model-3 & Concat & Diff  & 647.98 & 20.73 & 35.33 \\
    Model-4 & Diff  & Concat & 651.58 & 20.79 & 35.29 \\
    Model-5 & Concat & concat & 651.75 & 20.8  & 35.28 \\
    \textbf{Ours}  & Diff  & Diff  & 647.80 & 20.73 & \textbf{35.38} \\
\hline
    \end{tabular}%
  \label{abla-2}%
\end{table}%

\begin{table}[!t]
  \centering
  \captionsetup{font={scriptsize}}   
  \caption{Ablation experiment on the effectiveness of bidirectional difference. Results are calculated on scene 2 of Jilin-T in the Y channel.}
\renewcommand\arraystretch{1.2}  
    \begin{tabular}{c|cc|ccc}
\hline
    Models & Forward & Backward & FLOPs & \#Param. & PSNR (dB) \\
\hline
    Model-6 & \checkmark     &       & 636.23 & 20.73 & 35.15 \\
    Model-7 &       & \checkmark     & 636.23 & 20.73 & 35.28 \\
    \textbf{Ours}  & \checkmark     & \checkmark     & 647.80 & 20.73 & \textbf{35.38} \\
\hline
    \end{tabular}%
  \label{abla-3}%
\end{table}%
\subsubsection{Effectiveness of Difference Operation}
To investigate the effectiveness and efficiency of employing difference calculations in TDMs, we substitute the difference operation with naive concatenation. Specifically, in Model-3, we directly concatenate adjacent frames within S-TDM to extract local temporal information. Model-4 stacks the forward and backward features in the channel dimension and subsequently fuses them by a $3\times3$ convolution layer to extract the long-range dependency. Model-5 involves simultaneous concatenation within both S-TDM and L-TDM. The training process is shown in Fig. \ref{training}, and the PSNR results are reported in Table. \ref{abla-2}. It can be seen that simple concatenation results in a PSNR decrease of 0.1dB, accompanied by increased FLOPs (3.95G) and parameters (0.07M). Conversely, our LGTD can improve the efficiency and boost PSNR by conducting temporal difference learning TDMs, which demonstrates difference operation can reduce the high redundancy within satellite video and give an accurate representation of motion information.

\subsubsection{Effectiveness of Bidirectional Difference}
We argue that our L-TDM can leverage the long-range temporal information from forward and backward segments. Therefore, we establish two models for comparisons: (1) Model-6 with only forward propagation; (2) Model-7 with only backward propagation. The PNSR results are reported in Table. \ref{abla-3}. In contrast to a single forward architecture, the bidirectional configuration yields a PSNR improvement of 0.23dB with only a marginal increment in FLOPs. On the one hand, it demonstrates that forward and backward features can provide mutually complementary information. On the other hand, it emphasizes our capability to capture cross-segment information through our bidirectional difference operation.

\subsubsection{Effectiveness of Difference Compensation Unit}
The restored HR target frame is required to maintain the spatial consistency of the LR target frame. Although S-TDM and L-TDM are capable of modeling temporal information for compensation, we still need to guide the S-TDM and L-TDM in focusing on valuable compensation while minimizing misalignment interference. In light of this, Model-8 is introduced. In this model, the Difference Compensation Unit (DCU) is omitted, and instead, the temporal compensated feature is directly concatenated with the spatial feature for spatial-temporal fusion. The results, as listed in Table. \ref{abla-4},  indicate that DCU can explore more useful temporal information to improve spatial fidelity.

\subsubsection{Effectiveness of Hybrid Attention Reconstruction}
In our LGTD, two potent attention mechanisms are incorporated for the final reconstruction. We establish a baseline model that employs an equivalent count of residual blocks for reconstruction. However, due to the entanglement of short-term and long-term temporal information in the compensated feature, traditional residual blocks face challenges in handling such complex distributions. Model-10, which exclusively employs long-term attention, and Model-11, which solely focuses on short-term information, both encounter limitations by only partially addressing local and global redundancies. In summary, our hybrid attention emerges as a successful strategy for reconstructing temporal information for better restoration.



\begin{table}[!t]
  \centering
  \captionsetup{font={scriptsize}}   
  \caption{Ablation experiment on the effectiveness of difference compensation unit. Results are calculated in scene-2 of Jilin-T in the Y channel.}
\renewcommand\arraystretch{1.2}  
    \begin{tabular}{c|cc|ccc}
\hline
    Models & DF & DCU   & FLOPs & \#Param. & PSNR (dB) \\
\hline
    Model-8 & \checkmark     &       & \multicolumn{1}{r}{641.18} & 20.47 & 35.31 \\
    \textbf{Ours}  &       & \checkmark     & 647.80 & 20.73 & \textbf{35.38} \\
\hline
    \end{tabular}%
  \label{abla-4}%
\end{table}%

\begin{table}[!t]
  \centering
  \captionsetup{font={scriptsize}}   
  \caption{Ablation experiment on the effectiveness of hybrid attention reconstruction. Results are calculated in scene-2 of Jilin-T in the Y channel.}
\renewcommand\arraystretch{1.2}  
    \begin{tabular}{c|c|c|c|ccc}
\hline
    Models & Res-block & LA    & SA     & \#Param. & PSNR (dB) \\

\hline
    Model-9 & \checkmark     &       &        & 6.88 & 34.83 \\
    Model-10 &       & \checkmark     &        & 15.15 & 35.26 \\
    Model-11 &       &       & \checkmark      & 13.06 & 34.72 \\
    \textbf{Ours}  &       & \checkmark     & \checkmark      & 20.73 & \textbf{35.38} \\
\hline
    \end{tabular}%
  \label{abla-5}%
\end{table}%


\section{Conclusion}\label{conclu}
In this paper, we propose a Local-Global Temporal Difference learning network (LGTD) to realize the short-term and long-term temporal compensation for satellite VSR. To explore the local temporal information, we proposed to compute the frame-wise RGB difference between adjacent frames and supply short-term temporal representation to the target feature in a two-stage manner. To exploit the long-range dependency in the entire frame sequence, we design a bidirectional long-term temporal difference modeling branch. The holistic temporal difference is conducted on the smoothed forward and backward features to reduce the high redundancy. Then, the activated long-term attention is used to modulate the smoothed feature for compensation. Furthermore, we devise a difference compensation unit (DCU) to mitigate the misalignment of temporal difference learning. Extensive experiments on five mainstream satellite videos demonstrate our LGTD can recover high fidelity compared with state-of-the-art methods.

Although the proposed TDMs could deliver favorable temporal compensation, reconstructing high-quality target frames from local and global temporal compensated features remains a challenging problem. This paper employed a hybrid attention mechanism for the final reconstruction, which combines self-attention and channel attention for effective restoration. In spite of achieving decent performance, the hybrid attention increases the number of parameters dramatically. In future work, we plan to design a lightweight attention mechanism to realize the final reconstruction. This will allow us to maintain the capability of reconstructing local and global compensation while reducing the model size for more practical deployment.

\section{Acknowledgment}
We gratefully acknowledge the foundational work of the TDN framework \cite{wang2021tdn}, which provided important technical insights for the design of our method. Their contribution laid a solid basis for our temporal modeling component.



%
%



%

%


%
%

\ifCLASSOPTIONcaptionsoff
  \newpage
\fi



%
\bibliographystyle{IEEEtran}
\bibliography{reference}


%

\begin{IEEEbiography}[{\includegraphics[width=1in,height=1.25in,clip,keepaspectratio]{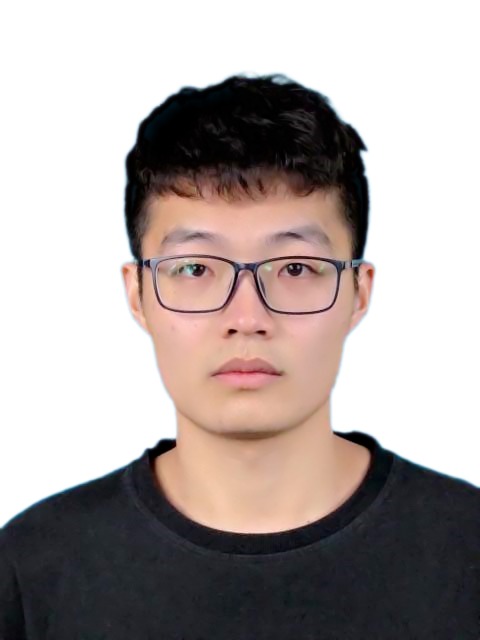}}]{Yi Xiao}
received the B.S. degree from the School of Mathematics and Physics, China University of Geosciences, Wuhan, China, in 2020. He is pursuing the Ph.D. degree with the School of Geodesy and Geomatics, Wuhan University, Wuhan.
\par His major research interests are remote sensing image super-resolution and computer vision. More details can be found at \url{https://xy-boy.github.io/}
\end{IEEEbiography}

\begin{IEEEbiography}[{\includegraphics[width=1in,height=1.25in,clip,keepaspectratio]{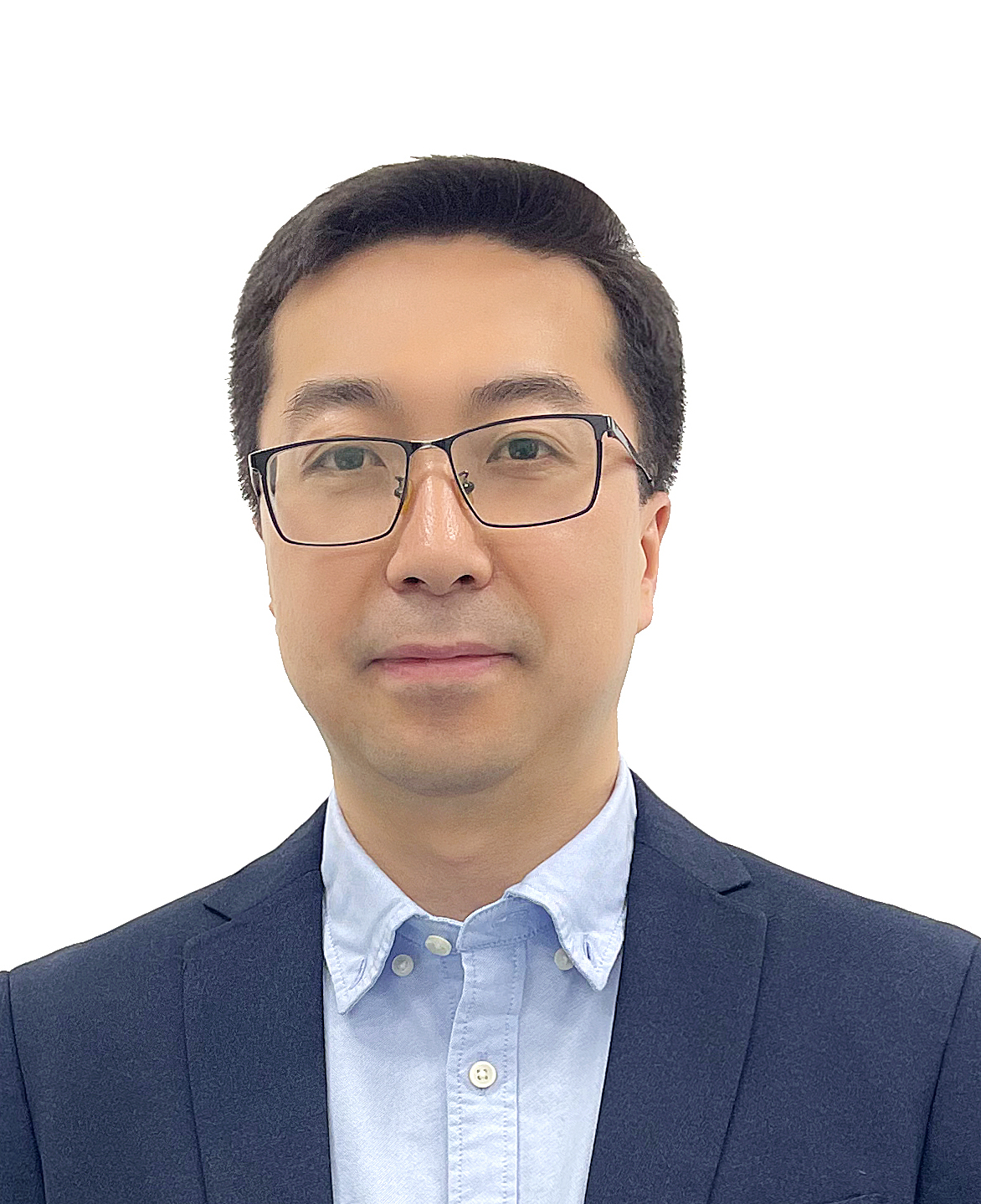}}]{Qiangqiang Yuan}
(Member, IEEE) received the B.S. degree in surveying and mapping engineering and the Ph.D. degree in photogrammetry and remote sensing from Wuhan University, Wuhan, China, in 2006 and 2012, respectively.
\par In 2012, he joined the School of Geodesy and Geomatics, Wuhan University, where he is a Professor. He has published more than 90 research papers, including more than 70 peer-reviewed articles in international journals, such as \emph{Remote Sensing of Environment, ISPRS Journal of Photogrammetry and Remote Sensing}, {\sc IEEE Transaction ON Image Processing}, and {\sc IEEE Transactions ON Geoscience AND Remote Sensing}. His research interests include image reconstruction, remote sensing image processing and application, and data fusion.
\par Dr. Yuan was a recipient of the Youth Talent Support Program of China in 2019, the Top-Ten Academic Star of Wuhan University in 2011, and the recognition of Best Reviewers of the IEEE GRSL in 2019. In 2014, he received the Hong Kong Scholar Award from the Society of Hong Kong Scholars and the China National Postdoctoral Council. He is an associate editor of 5 international journals and has frequently served as a referee for more than 40 international journals for remote sensing and image processing.
\end{IEEEbiography}

\begin{IEEEbiography}[{\includegraphics[width=1in,height=1.25in,clip,keepaspectratio]{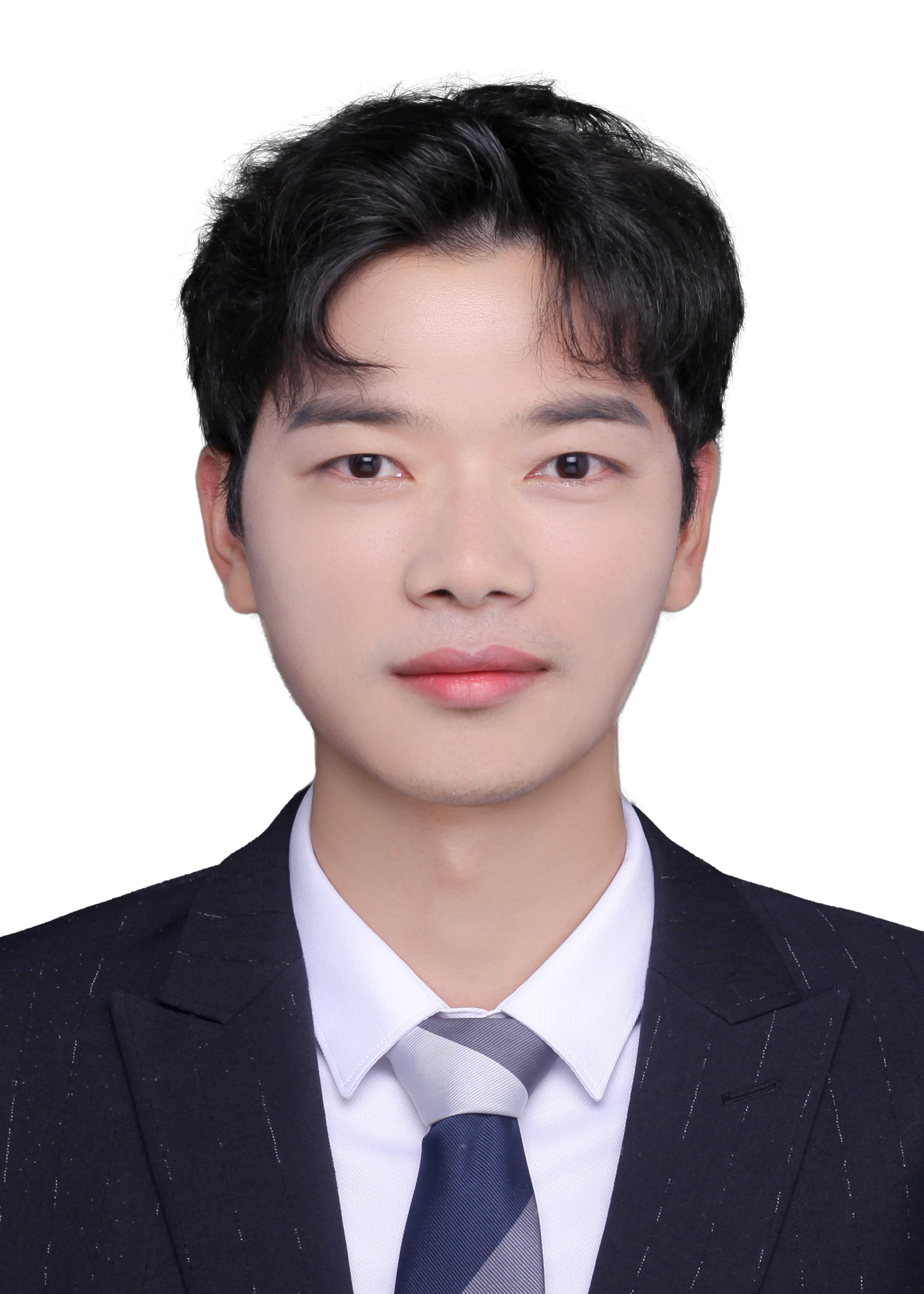}}]{Kui Jiang}
(Member, IEEE) received the Ph.D. degree in the School of Computer Science, Wuhan University, Wuhan, China, in 2022. Dr. Jiang is now a Chief Engineer with Huawei Technologies, Cloud BU, Hangzhou, China. He received the  2022 ACM Wuhan Doctoral Dissertation Award. 
\par His research interests include image/video processing and computer vision.
\end{IEEEbiography}

\begin{IEEEbiography}[{\includegraphics[width=1in,height=1.25in,clip,keepaspectratio]{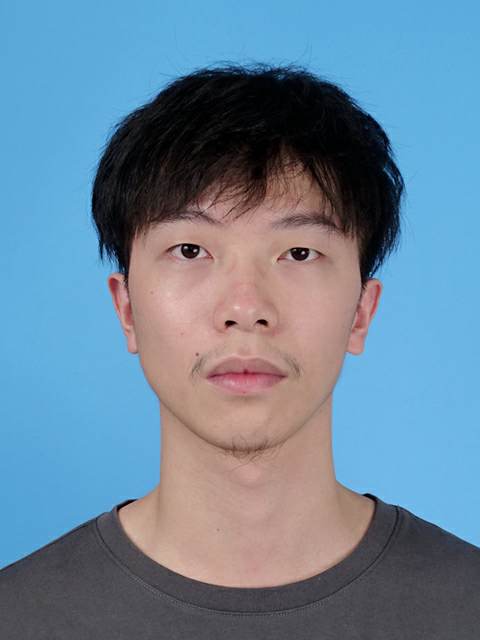}}]{Xianyu Jin}
received the B.S. degree in geodesy and geomatics engineering from Wuhan University, Wuhan, China, in 2019, where he is pursuing the M.S. degree with the School of Geodesy and Geomatics.
\par His research interests include video super-resolution, deep learning, and computer vision.
\end{IEEEbiography}

\begin{IEEEbiography}[{\includegraphics[width=1in,height=1.25in,clip,keepaspectratio]{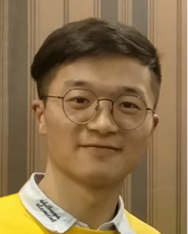}}]{Jiang He}
received the B.S. degree in remote sensing science and technology from faculty of geosciences and environmental engineering in Southwest Jiaotong University, Chengdu, China, in 2018. He is currently pursuing the Ph.D. degree in School of Geodesy and Geomatics, Wuhan University, Wuhan, China.
\par His research interests include hyperspectral super-resolution, image fusion, quality improvement, remote sensing image processing and deep learning.
\end{IEEEbiography}

\begin{IEEEbiography}[{\includegraphics[width=1in,height=1.25in,clip,keepaspectratio]{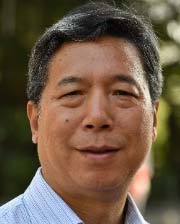}}]{Liangpei Zhang}
(Fellow, IEEE) received the B.S. degree in physics from Hunan Normal University, Changsha, China, in 1982, the M.S. degree in optics from the Xi’an Institute of Optics and Precision Mechanics, Chinese Academy of Sciences, Xi’an, China, in 1988, and the Ph.D. degree in photogrammetry and remote sensing from Wuhan University, Wuhan, China, in 1998.
\par He is currently a “Chang-Jiang Scholar” Chair Professor appointed by the Ministry of Education of China at the State Key Laboratory of Information Engineering in Surveying, Mapping, and Remote Sensing (LIESMARS), Wuhan University. He was a Principal Scientist for the China State Key Basic Research Project from 2011 to 2016 appointed by the Ministry of National Science and Technology of China to lead the Remote Sensing Program in China. He has published more than 700 research articles and five books. He is the Institute for Scientific Information (ISI) Highly Cited Author. He holds 30 patents. His research interests include hyperspectral remote sensing, high-resolution remote sensing, image processing, and artificial intelligence. 
\par Dr. Zhang is a fellow of the Institution of Engineering and Technology (IET). He was a recipient of the 2010 Best Paper Boeing Award, the 2013 Best Paper ERDAS Award from the American Society of Photogrammetry and Remote Sensing (ASPRS), and the 2016 Best Paper Theoretical Innovation Award from the International Society for Optics and Photonics (SPIE). His research teams won the top three prizes of the IEEE GRSS 2014 Data Fusion Contest. His students have been selected as the winners or finalists of the IEEE International Geoscience and Remote Sensing Symposium (IGARSS) Student Paper Contest in recent years. He is also the Founding Chair of the IEEE Geoscience and Remote Sensing Society (GRSS) Wuhan Chapter. He also serves as an associate editor or an editor for more than ten international journals. He is also serving as an Associate Editor for the {\sc IEEE Transactions ON Geoscience AND Remote Sensing}.
\end{IEEEbiography}

\begin{IEEEbiography}[{\includegraphics[width=1in,height=1.25in,clip,keepaspectratio]{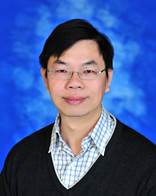}}]{Chia-Wen Lin}
(Fellow, IEEE) received the Ph.D degree in electrical engineering from National Tsing Hua University (NTHU), Hsinchu, Taiwan, in 2000. He is currently a Professor at the Department of Electrical Engineering and the Institute of Communications Engineering at NTHU. He is also Deputy Director of the AI Research Center of NTHU. He was with the Department of Computer Science and Information Engineering, National Chung Cheng University, Taiwan, during 2000–2007. Prior to joining academia, he worked for the Information and Communications Research Laboratories, Industrial Technology Research Institute, Hsinchu, Taiwan, during 1992– 2000. His research interests include image and video processing, computer vision, and video networking. 
\par Dr. Lin served as a Distinguished Lecturer of IEEE Circuits and Systems Society from 2018 to 2019, a Steering Committee member of IEEE TRANSACTIONS ON MULTIMEDIA from 2014 to 2015, and the Chair of the Multimedia Systems and Applications Technical Committee of the IEEE Circuits and Systems Society from 2013 to 2015. His articles received the Best Paper Award of IEEE VCIP 2015 and the Young Investigator Award of VCIP 2005. He received the Outstanding Electrical Professor Award presented by the Chinese Institute of Electrical Engineering in 2019, and the Young Investigator Award presented by the Ministry of Science and Technology, Taiwan, in 2006. He is also the Chair of the Steering Committee of IEEE ICME. He has served as a Technical Program Co-Chair for IEEE ICME 2010, and a General Co-Chair for IEEE VCIP 2018, and a Technical Program Co-Chair for IEEE ICIP 2019. He has served as an Associate Editor of IEEE TRANSACTIONS ON IMAGE PROCESSING, IEEE TRANSACTIONS ON CIRCUITS AND SYSTEMS FOR VIDEO TECHNOLOGY, IEEE TRANSACTIONS ON MULTIMEDIA, IEEE MULTIMEDIA, and Journal of Visual Communication and Image Representation.
\end{IEEEbiography}





\end{document}